\PassOptionsToPackage{dvipsnames,table}{xcolor}
\documentclass[sigconf=true, nonacm=true, review=false, anonymous = false]{acmart}
\usepackage{booktabs} 
\usepackage{amsmath}
\usepackage{xspace}
\usepackage{xcolor}
\usepackage{dsfont}
\usepackage{rotating}
\usepackage{makecell}

\usepackage{amsthm}
\usepackage{subcaption}
\renewcommand{\epsilon}{\varepsilon}

\usepackage{mathtools}
\DeclarePairedDelimiter{\ceil}{\lceil}{\rceil}

\newcommand{\R}{\mathbb{R}}

\newcommand{\hq}{\textbf{high-quality modCMA configurations}\xspace}
\newcommand{\lowq}{\textbf{random modCMA configurations}\xspace}

\newcommand{\AUC}{\ensuremath{\operatorname{AUC}}\xspace}
\newcommand{\AOC}{\ensuremath{\operatorname{AOC}}\xspace}

\begin{document}

\title{Analyzing the Impact of Undersampling on the Benchmarking and
  Configuration of Evolutionary Algorithms}%
\thanks{Please cite as: Diederick Vermetten, Hao Wang, Manuel López-Ibañez,
  Carola Doerr, and Thomas Bäck. 2022. Analyzing the Impact of Undersampling on
  the Benchmarking and Configuration of Evolutionary Algorithms. In
  \emph{Genetic and Evolutionary Computation Conference (GECCO ’22)}, July
  9–13, 2022, Boston, MA, USA. ACM, New York, NY, USA, 9
  pages. \url{https://doi.org/10.1145/3512290.3528799}}

\author{Diederick Vermetten}
\orcid{0000-0003-3040-7162}
\affiliation{
  \institution{Leiden Institute for Advanced Computer Science}
  \city{Leiden}
  \country{The Netherlands}
}
\email{d.l.vermetten@liacs.leidenuniv.nl}

\author{Hao Wang}
\orcid{0000-0002-4933-5181}
\affiliation{
  \institution{Leiden Institute for Advanced Computer Science}
  \city{Leiden}
  \country{The Netherlands}
}
\email{h.wang@liacs.leidenuniv.nl}

\author{Manuel L\'opez-Iba\~nez}
\orcid{0000-0001-9974-1295}
\affiliation{
  \institution{School of Computer Science and Engineering, University of M\'alaga}
  \city{M\'alaga}
  \country{Spain}
}
\email{manuel.lopez-ibanez@uma.es}
\author{Carola Doerr}
\orcid{0000-0002-4981-3227}
\affiliation{
  \institution{Sorbonne Universit\'e, CNRS, LIP6}
  \city{Paris}
  \country{France}}
\email{Carola.Doerr@lip6.fr}

\author{Thomas B{\"a}ck}
\orcid{0000-0001-6768-1478}
\affiliation{
  \institution{Leiden Institute for Advanced Computer Science}
  \city{Leiden}
  \country{The Netherlands}
}
\email{t.h.w.back@liacs.leidenuniv.nl}

\begin{abstract}
The stochastic nature of iterative optimization heuristics leads to inherently noisy performance measurements. Since these measurements are often gathered once and then used repeatedly, the number of collected samples will have a significant impact on the reliability of algorithm comparisons. 
We show that care should be taken when making decisions based on limited data. 
Particularly, we show that the number of runs used in many benchmarking studies, e.g., the default value of 15 suggested by the COCO environment, can be insufficient to reliably rank algorithms on well-known numerical optimization benchmarks.

Additionally, methods for automated algorithm
configuration are sensitive to insufficient sample sizes. 
This may result in the configurator choosing a ``lucky'' but poor-performing configuration despite exploring better ones.  We show that relying on mean performance values, as many configurators do, can require a large number of runs to provide accurate comparisons between the considered configurations. Common statistical tests can greatly improve the situation in most cases but not always. 
We show examples of performance losses of more than $20\%$, even when using statistical races to dynamically adjust the number of runs, as done by irace.
Our results underline the importance of appropriately considering the statistical distribution of performance values.

\end{abstract}



\maketitle

\section{Introduction}

The study of iterative optimization heuristics is a continuously developing area within computer science. With the ever expanding number of newly developed algorithms, the importance of proper benchmarking has been gaining more traction~\cite{BarDoeBer2020benchmarking}. Standardized benchmarking environments have been proposed for a wide variety of problem types~\cite{HanAugMer2020coco,IOHprofiler,nevergrad} and are widely used to compare the performance of different optimizers. However, since most state-of-the-art optimizers are stochastic in nature, assessing their performance is not necessarily a straightforward task, as any selected measure for performance is inherently the result of a limited sampling of some underlying probability distribution. 

In order to get reliable estimates for the distributions of these performance measures, benchmarking pipelines generally recommend to perform multiple optimization runs on each problem. For example, the popular COCO environment~\cite{HanAugMer2020coco} recommends measuring the performance of 15 independent runs. When comparing algorithms, these individual performance measures are aggregated into a single number per function, using a variety of different measures ranging from taking the mean of hitting times, to Expected Running Time (ERT) or anytime performance metrics such as the area under the Empirical Cumulative Distribution Function (ECDF). 

 Aggregated performance values calculated from a limited number of runs may poorly estimate the true expected performance of an algorithm, especially when the distribution of individual performance values is non-normal or shows a large variance. It is clear that measuring only one run will lead to many mistakes when comparing a set of algorithms in an algorithm selection or configuration context, but it is not clear whether 15, 50 or more runs are sufficient to alleviate this issue.

In addition to comparing aggregated values such as the mean, statistical tests are often used to decide whether one algorithm outperforms another. Widely-used examples are the parametric t-test, when distributions are assumed to be somewhat normal, or the non-parametric Wilcoxon rank-sum test, when this assumption cannot be made. Although the hypothesis tested by the Wilcoxon rank-sum test refers to the relative ranking of independent samples and cannot be used to conclude anything about mean values, this distinction is often ignored in the literature when drawing conclusions from it. 

The problem of comparing algorithms based on a potentially small number of samples is not limited to benchmarking studies, but also applies to meta-optimization, such as algorithm configuration~\cite{HuaLiYao2020algconf}, also known as automated tuning, or algorithm selection~\cite{KerHooNeuTra2019}. When faced with an algorithm configuration problem, the configurator has to make a tradeoff between collecting more samples from promising configurations to gain confidence in their performance and exploring a larger variety of configurations. Most commonly, these samples are then compared based on their mean or using statistical tests to determine which configurations are returned as the best-found. 

Previous work~\cite{VerWanDoeBac2020cash} has shown that configuration methods are susceptible to large performance variance, which may lead to a potential loss of performance with respect to the true best configurations explored.

In this work, we highlight the challenges inherent in comparing the performance of stochastic optimization algorithms. We show that, for several cases, 
the currently recommended values for the number of samples and testing procedure can lead to mistakes. We show that the distribution of performance values has a large impact on algorithm configuration methods, indicating that there is not one method of performance comparison that dominates all others. 
Importantly, our results demonstrate that we must identify better ways to handle the stochasticity of iterative optimization heuristics when applying algorithm configuration methods. 

\section{Preliminaries}\label{sec:prelim}

\subsection{Performance Measures}\label{sec:perf_measure}
We consider the minimization of functions of the form: \[ f\colon X\subset\R^d\to \R \]

where $X$ is the \emph{search space} and $d$ denotes its dimensionality. We focus on iterative optimization heuristics (IOHs), and in particular randomized IOHs. IOHs optimize $f$ by a sequential process of generating solution candidates $x^1, \dotsc, x^{\lambda} \in X$, evaluating their quality $f(x^1), \dotsc, f(x^{\lambda})$, and adjusting their strategy to generate the next candidates. For randomized IOHs, both the \emph{number} of candidates that are generated in a given iteration and the \emph{strategy} to generate them can be stochastic.  

While there are a large number of metrics that could be considered to measure the performance of these algorithms, we limit ourselves to \emph{Area Under the ECDF Curve (AUC)}. AUC is an anytime performance measure defined as follows.

\begin{definition}[Area Under the ECDF Curve, AUC]

For a given optimization algorithm $A$ with a budget of $B$ function evaluations for minimizing a function $f\colon X \to \R$ and a given finite set of targets $\mathcal{V} \subset \R$, the AUC value of $A$ on $f$ is approximated by  
\begin{align}
    \AUC(A, f, \mathcal{V}) &= \int_1^B \widehat{F}(t; A,f,\mathcal{V}) \mathrm{d} t\enspace,
\end{align}
where 
\begin{align}
    \widehat{F}(t; A,f, \mathcal{V}) &= \frac{1}{N|\mathcal{V}|}\sum_{\phi\in\mathcal{V}}\sum_{i=1}^{N} \mathds{1}(t_i(A,f,\phi) \leq t)\enspace, 
\end{align}    
and $N$ is the number of (ideally independent) runs for which we have performance logs, and $\mathds{1}(t_i(A,f,\phi)\leq t)$ is an indicator function that returns 1 if the first hitting time of target $\phi$ in run $i$ of $A$ is not larger than $t$. If the target $\phi$ is not hit in this run, the indicator always returns $0$. 
\end{definition}

AUC is a measure that should be maximized, however, most automated algorithm configuration methods are designed for minimization. Thus, we use the \emph{Area Over the Curve (AOC)} instead of AUC. Since the values $\widehat{F}(t; A,f, \mathcal{V})$ are normalized between 0 and 1, the AUC value is a real number between 0 and $B$. We can hence define $\AOC(A, f, \mathcal{V})=B-\AUC(A, f, \mathcal{V})$. 
For our experiments, we set the set of targets $\mathcal{V}$ to be the default as used in the COCO environment (51 targets, logarithmically spaced between $10^2$ and $10^{-8}$).

\subsection{Benchmark Functions}\label{sec:bbob}
For this study, we make use of the single-objective, noiseless functions from COCO's BBOB suite~\cite{HanAugMer2020coco}, which is considered to be one of the most popular sets of benchmark functions for benchmarking continuous derivative-free black-box optimization algorithms. For our experiments, we use the $5$-dimensional versions of these functions. 

Benchmarking studies using BBOB commonly perform each individual run on a different  ``instance'' of each function, where each instance is generated by applying transformations in both the domain and objective space, in such a way that the core properties of the function are preserved~\cite{HanFinRosAug2009bbob}. In this work we aim to avoid the additional variance caused by multiple instances of each function, and thus focus on a single instance (instance ID 1) of each of the 24 BBOB functions.

\subsection{Algorithm Configuration: Irace}\label{sec:irace}
As part of this work, we investigate the impact of performance variability on algorithm configuration. To perform algorithm configuration, we use irace~\cite{LopDubPerStuBir2016irace}, which is based on an iterated racing procedure. Irace starts out by randomly generating a set of parameter configurations using uniform sampling in the defined parameter space. Then, the first race starts, where for each of these configurations, a number of runs (\textit{FirstTest}) are performed, after which a statistical test (either t-test or Friedman test) determines which configurations to continue with. The surviving configurations are run again a certain number of times (\textit{EachTest}, set to 1 in this paper). Then the test is performed again to potentially eliminate additional configurations, and runs keep being added in this way until no more than a given number ($5$ in this paper) of configurations remain (elites)

or the budget assigned to this race is exhausted. The elite configurations are then used as the basis for generating new candidates to be evaluated in the next race, until the overall budget for irace is fully used. This procedure eventually leads to $5$ or fewer surviving elite configurations, from which we select the one with the lowest mean as the final recommendation.

This racing approach is however not the only technique used by algorithm configurators to select the best configurations from a larger set.  Some algorithm configuration methods, such as Hyperband~\cite{LiJamSal2018hyperband}, use a successive halving (SHA) approach~\cite{KarKorSom2013}. In the first step of SHA, \textit{FirstTest} runs are performed for each of the $n$ initial configurations. Then, given a reduction factor $R$, the $\ceil{n/R}$ configurations with the lowest mean survive, and the others are discarded. For the surviving configurations, an additional $2\cdot\textit{FirstTest}$ runs are performed. This process of keeping the best $1/R$ fraction based on mean, and doubling the number of additional runs, is repeated until only one configuration survives.

\begin{figure*}[!t]
    \centering
    \includegraphics[width=\textwidth, trim=0mm 7mm 0mm 0mm, clip]{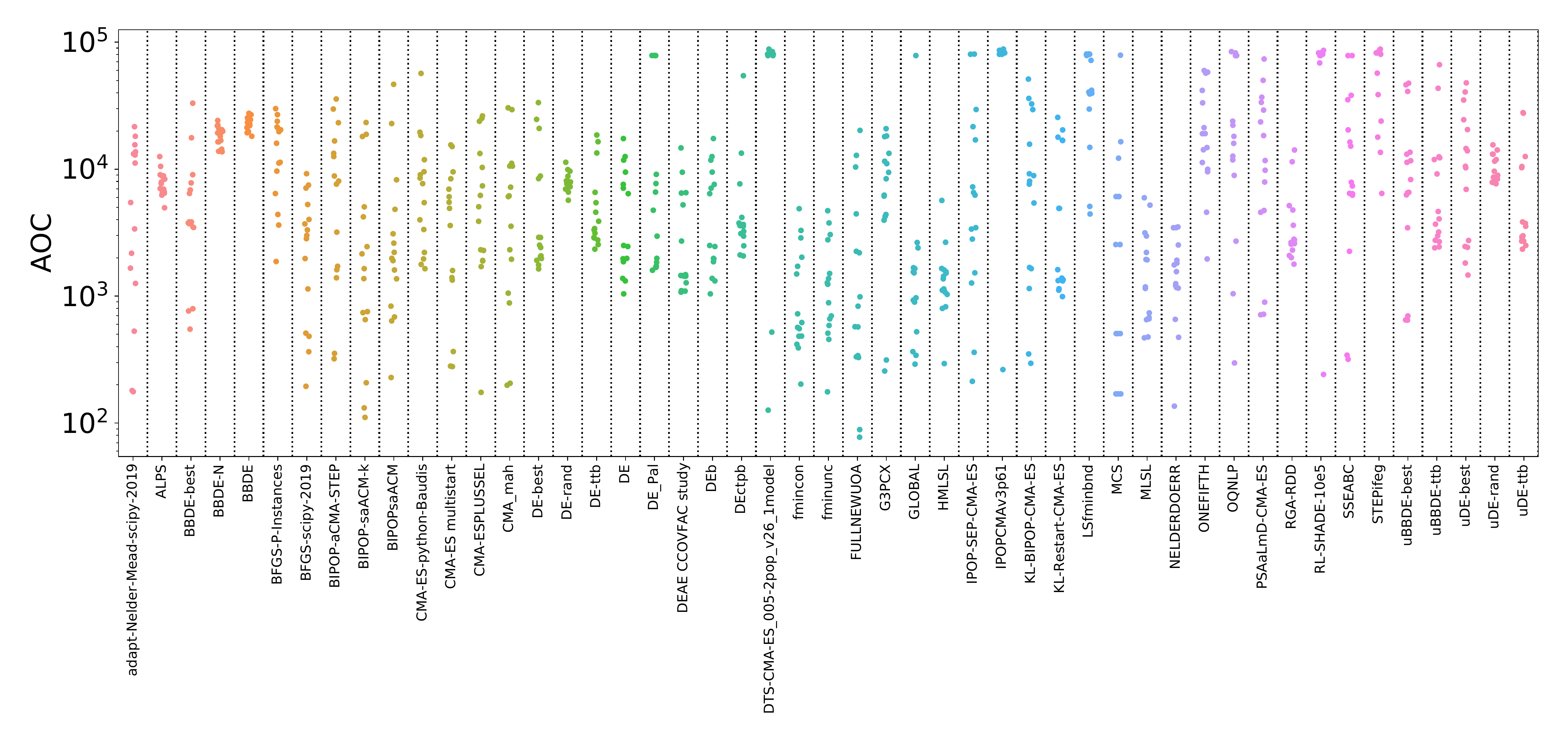}
    \caption{Distribution of the \AOC values of 15 independent runs of available BBOB algorithms on F21 in 5D.}
    \label{fig:F21_Aoc_distr}
\end{figure*}

\subsection{Modular CMA-ES}\label{sec:modcma}
To study the impact of performance variability on algorithm configuration, we make use of the Modular CMA-ES (modCMA) 
framework~\cite{modCMA,NobVerWan2021gecco}. This framework provides an implementation of the popular CMA-ES~\cite{HanOst1996cma} algorithm, with a wide variety of different modifications that can be activated independently from each other. In this work, we consider a hyper-parameter space consisting of $10$ discrete modules and 4 continuous hyper-parameters. This search space corresponds to the baseline used to analyze the modular CMA-ES in~\cite{NobVerWan2021gecco}, and we thus base our initial analysis on the same data~\cite{NobVerWan2021gecco-supp} to avoid needlessly re-running irace. The data we use consists of irace runs on each of the $24$ BBOB functions. Here, we use 1 of these irace runs for each function. Then, for each of the configurations sampled during these irace runs,
 we collect $200$ new independent runs, which we refer to as \emph{verification runs}.
Since the irace runs we use typically generated over 200 configurations each, this means we have collected at least $24\cdot 200 \cdot 200 =960\,000$ runs that we analyze in this work. We also use this data to simulate independent races. 

Throughout this paper, we will refer to two sets of configurations: the 33 configurations generated during the first race of irace are sampled uniformly at random in the configuration space, and are thus referred to as \lowq. Since we often want to consider the impact of the later parts of the tuning, we also consider the 33 last generated configurations. This set of configurations will be referred to as \hq.

\section{Why 15 runs are not enough}\label{sec:motivation}

While it is clear that any aggregated performance measure used to compare randomized algorithms is an empirical estimation of their true performance, the variance of this estimation is not necessarily equal for all algorithms on all functions. However, for a practical benchmarking setup, this nuance is often ignored in favour of simpler guidelines, such as aggregating a fixed number of \emph{samples} (i.e., individual performance values from independent runs) for each algorithm on each function.
The usual recommendation of 15 samples~\cite{HanAugMer2020coco} is often enough to make clear decisions on simple uni-modal functions, but the situation is much less clear on more challenging optimization problems. 

We illustrate the significant variation in performance between runs by showing in Figure~\ref{fig:F21_Aoc_distr} the distribution of 15 independent \AOC-values for a wide variety of algorithms from the BBOB-repository (\url{https://numbbo.github.io/data-archive/bbob/}) on F21 in 5D.
This figure also shows that the normality assumption, commonly taken for granted in benchmarking studies, is not well supported by the apparent distribution of the 15 performance values shown for each algorithm.

For some algorithms, the performance distributions even appear to show signs of bi-modality. As such, any analyses made based on this set of samples should be treated with care. While this large amount of variance is very pronounced in F21, it is not limited to this function, as other functions display similar effects but to a slightly lesser extent.\footnote{Plots equivalent to Figure~\ref{fig:F21_Aoc_distr} for other functions and performance measures are available on our figshare repository~\cite{figshare_figures}.}

The impact of performance variability can potentially be even larger when considering the task of algorithm configuration. It has previously been observed that the performance of an algorithm configuration on 

verification runs can differ significantly from the runs performed during the configuration task~\cite{NobVerWan2021gecco}. 

\begin{figure}
    \centering
    \includegraphics[width=0.48\textwidth]{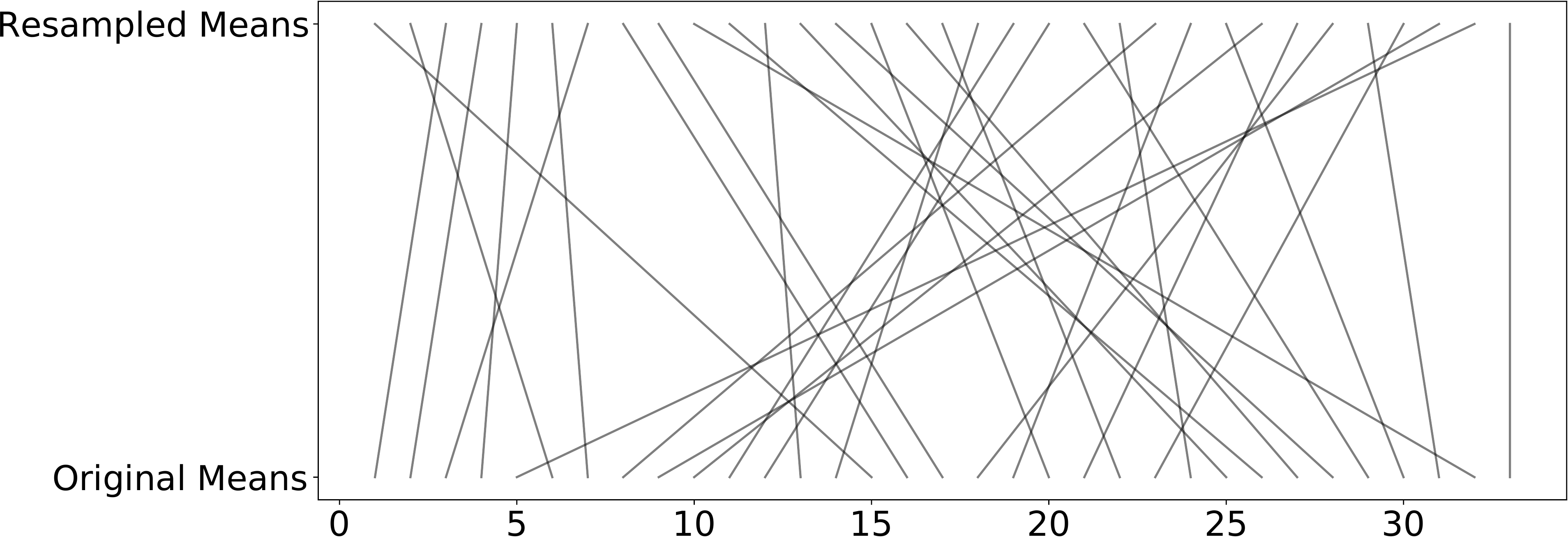}
    \caption{Changes in ranking of 33 \lowq based on calculating the mean over a sample of  15 AOC values (\emph{Resampled Means}) versus the 200 verification run samples (\emph{Original Means}), on F21.}
    \label{fig:example_rank_change}
\end{figure}

We illustrate this effect by showing in Figure~\ref{fig:example_rank_change} the changes in the ranking of the 33 \hq described in Section~\ref{sec:modcma} when calculating mean performance using a small sample size (15) and a larger number of verification runs (200) on F21.

\begin{figure}
    \centering
    \includegraphics[width=0.48\textwidth]{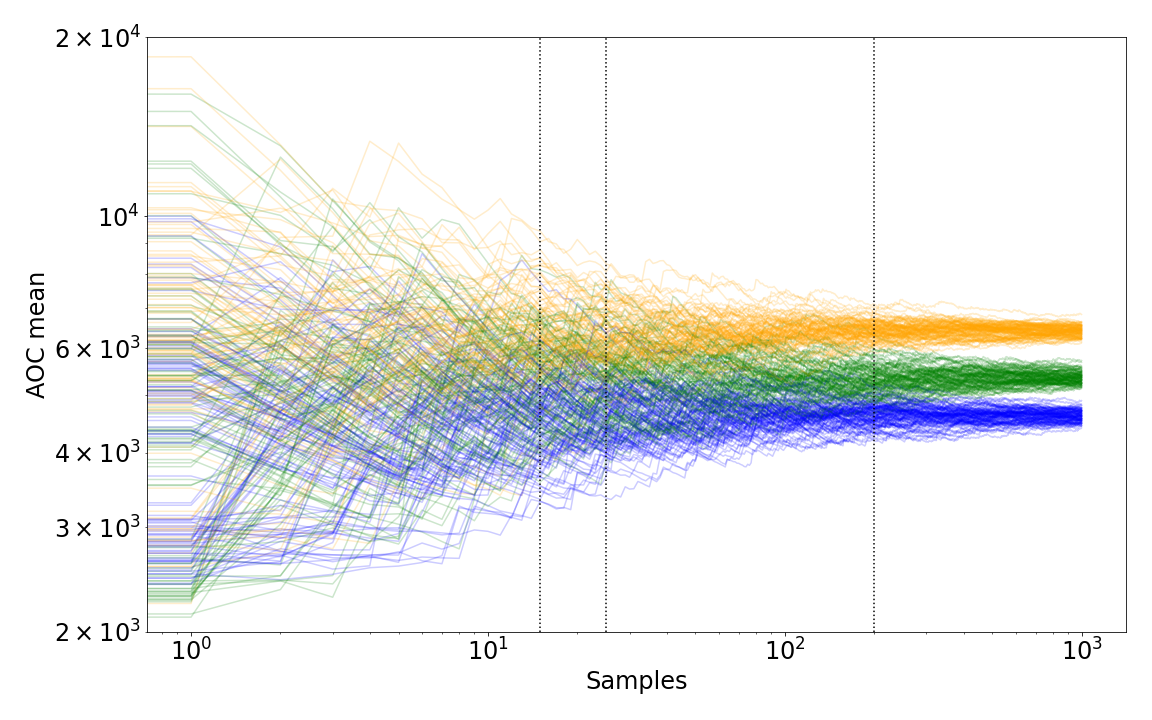}
    \caption{Evolution of the cumulative mean over sample sizes of 3 selected \hq on F18. The vertical lines indicate sample sizes 15, 25 and 200 respectively. Means are based on sampling with replacement from the original 200 samples of each configuration.}
    \label{fig:mean_over_time_modcma}
\end{figure}

While this might be considered a rather extreme case, it is by no means the only scenario in which behaviours like this can occur. Since algorithm configuration often generates similarly performing configurations near the end of a configuration run (while exploiting promising regions), making decisions about which configuration to select might become very noisy when using relatively low sample sizes. This phenomenon is exemplified in Figure~\ref{fig:mean_over_time_modcma}, where we show the evolution of the mean AOC of 3 selected \hq relative to an incremental number of AOC values. Each horizontal line of the same color corresponds to the cumulative mean of a sequence of values sampled with replacement from the same 200 AUC values. Despite sampling from the same 200 AUC values, the variance of the means of 15 and 25 samples is quite large and those means often poorly estimate the true mean performance.

In practice, making an incorrect decision between two configurations matters less when their true performance is very similar. However, when the set of configurations which are being compared increases in size, the risk of making incorrect decisions between more distinct configurations could potentially grow as well. In some situations in algorithm configuration tasks, we have observed significant differences between the performance of the selected elite configurations, and the best one from all configurations sampled according to the verification runs.

\begin{figure}[!tb]
    \centering
    \includegraphics[width=0.48\textwidth, trim={8mm, 10mm, 5mm, 0mm},clip]{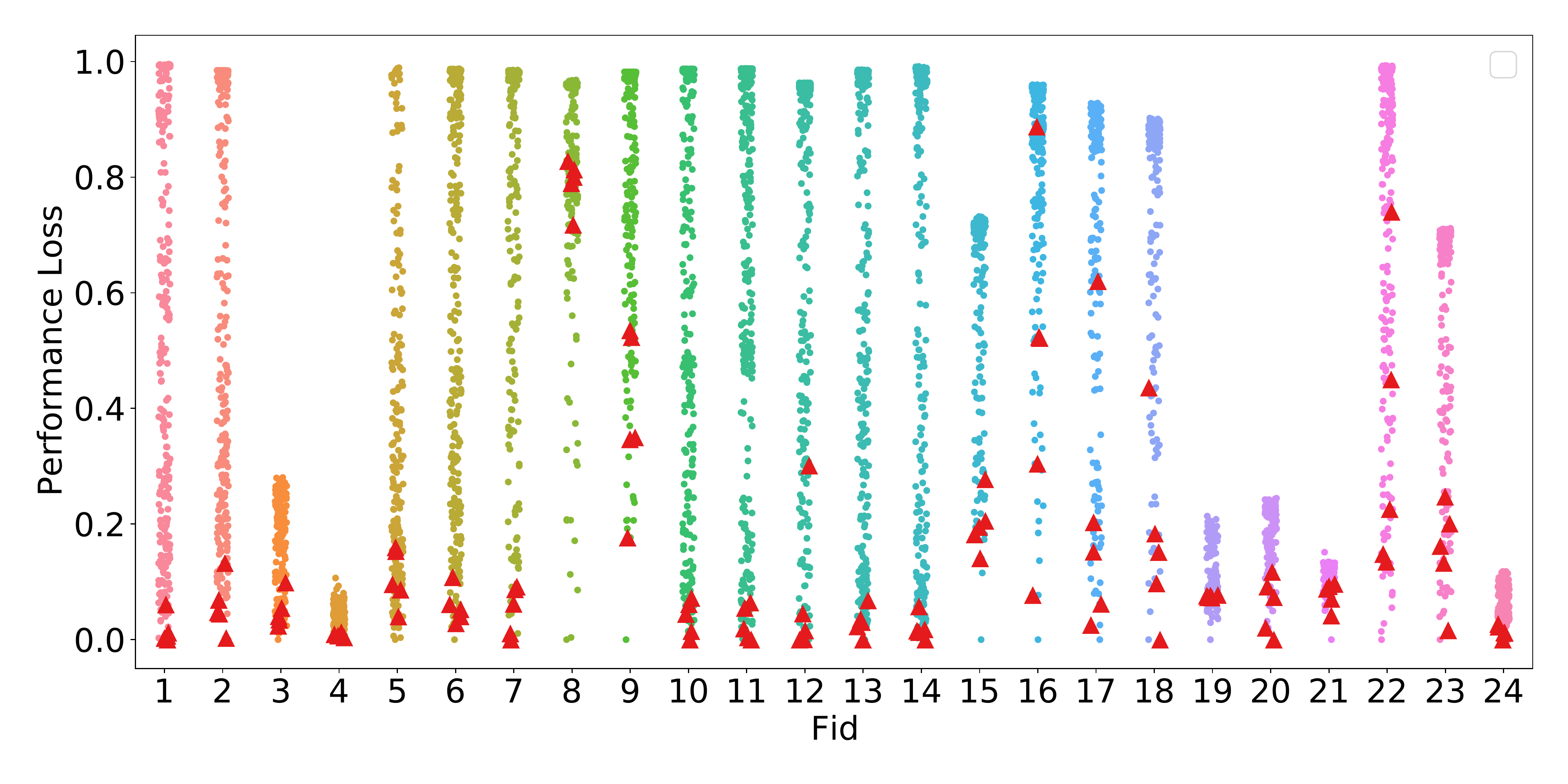}
    \caption{Performance losses between all modCMA configurations explored during the execution of one irace run on each function, and the best one from this set of configurations. The final elites of each irace run are marked in larger red triangles. All datapoints shown are based on 200 independent samples.}
    \label{fig:irace_perf_loss_all}
\end{figure}

In Figure~\ref{fig:irace_perf_loss_all}, we show the distribution of \AOC values for each configuration sampled during a run of irace on each of the 24 BBOB functions. The performance of each configuration is based on the mean of 200 verification runs, and the plot shows the relative performance loss to the best of these means. The (up to five) elite configurations returned by irace are marked with a red triangle. The lowest of these elites corresponds to the level of performance loss achieved by irace compared to the best-performing configuration sampled during the configuration process. From this figure, it can be seen that for some of the more complex functions, a $10\%$ performance loss or more can occur, clearly demonstrating that the variability of performance can severely hinder the outcome of the configuration efforts. 

\section{Impact on Benchmarking}\label{sec:benchmarking}

\begin{figure}
    \centering
    \includegraphics[width=0.48\textwidth, trim={8mm, 10mm, 5mm, 0mm},clip]{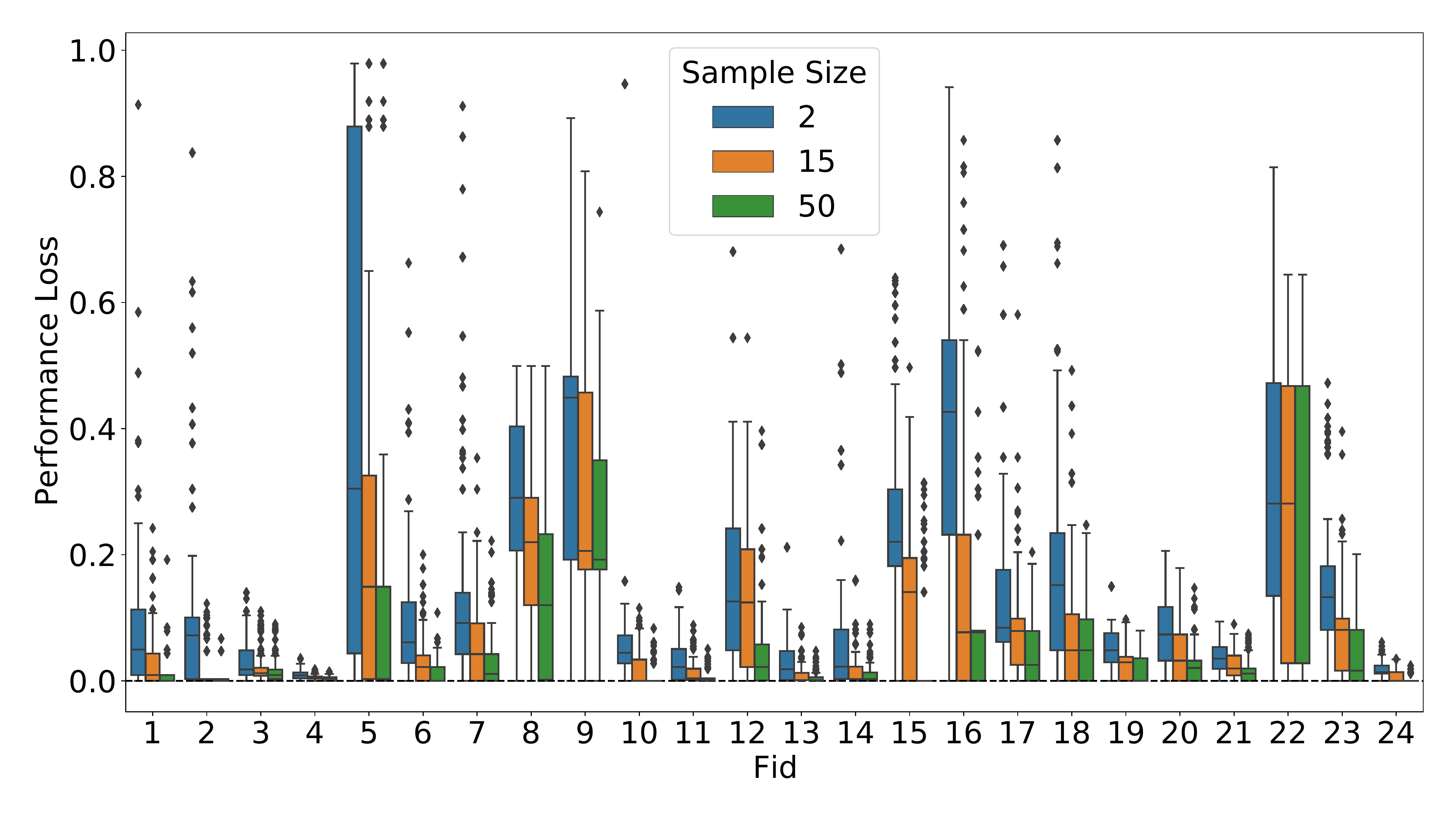}
    \caption{Performance loss, relative to the configuration with the best mean calculated over 200 samples, when comparing 33 \hq based on mean calculated from different number of samples. Each bar represents 5000 repetitions of the experiment.}
    \label{fig:perf_loss_mean_modcma}
\end{figure}

\begin{figure*}
    \centering
    \includegraphics[width=0.96\textwidth, trim={5mm, 10mm, 5mm, 0mm},clip]{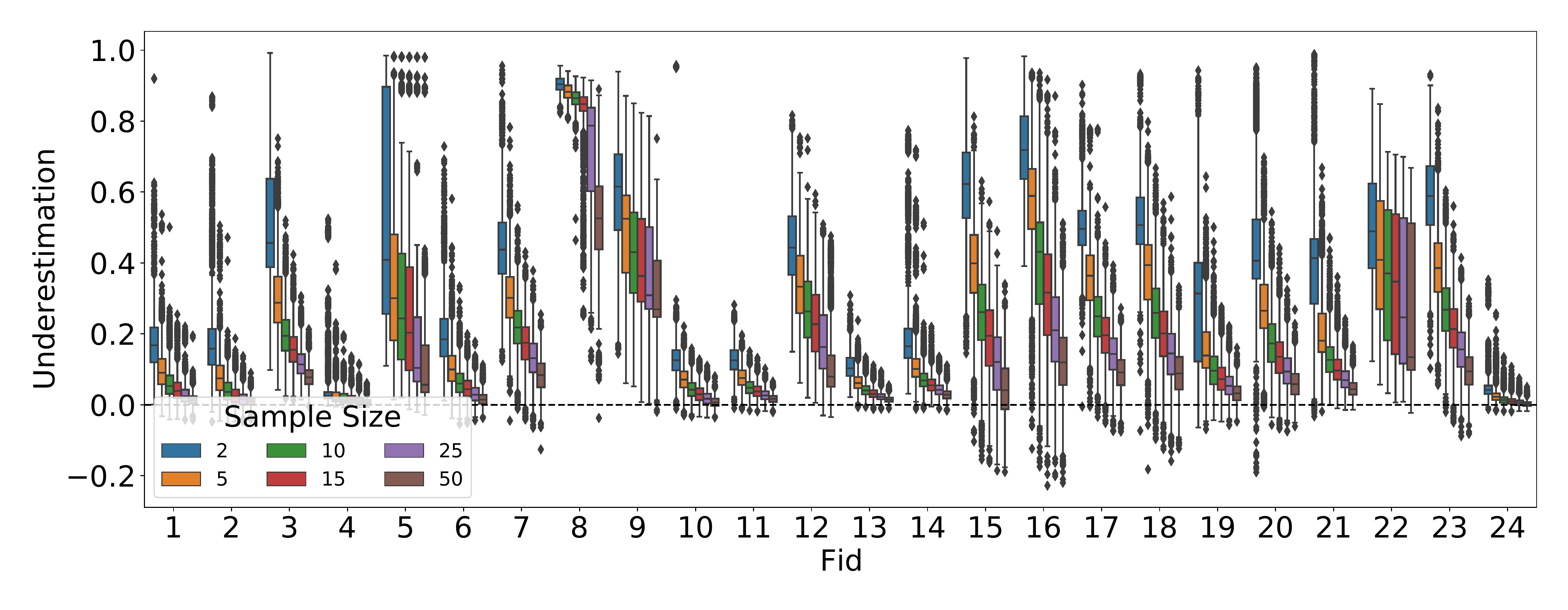}
    \caption{Underestimation when comparing 33 \hq based on mean calculated from different number of samples. Each bar represents 5000 repetitions of the experiment.
    }
    \label{fig:underestimation_mean_modcma}
\end{figure*}

To simulate a common algorithm comparison scenario, we make use of the set of 33 \hq from Section~\ref{sec:modcma} and simulate the benchmarking procedure by randomly re-sampling with replacement \AOC values, for sample sizes 2, 5, 10, 15, 25 and 50 from the set of 200. Then, we select the configuration with the best mean for each particular sample size as the winner, and compare its true performance (i.e., over 200 runs) to that of the actual best configuration to get an estimate for the performance loss. This process is repeated 5000 times for each sample size and each function, and the resulting performance loss per function is shown in Figure~\ref{fig:perf_loss_mean_modcma}.
We conclude that using means to determine the best-performing algorithm is not always reliable, and can lead to selecting configurations that are clearly sub-optimal. Many benchmarking studies use non-parametric statistical tests to assess significant differences without assuming normality, yet they still rely on comparison of means to rank the algorithms.
While we can see that increasing the used sample size is always beneficial, even using as many as 50 samples can still see performance losses of $10\%$ and more on some functions.

One possible explanation for these results is that, 
when we are determining the best algorithm from a large set of algorithms of wide performance variability, our decision is prone to underestimate the true mean due to the small sample size, i.e., we might ``luckily'' sample many good values for an algorithm with sub-optimal performance.

We quantify this impact by calculating, for each selected configuration and a given sample size, the \emph{underestimation error}, that is, the relative error of the mean estimated from the selected samples relative to its true mean performance (based on the 200 verification runs). Positive values indicate that the sample mean is lower, i.e., better, than the true mean.  We plot in Figure~\ref{fig:underestimation_mean_modcma} the underestimation error for \hq.

\begin{figure}
    \centering
    \includegraphics[width=0.48\textwidth]{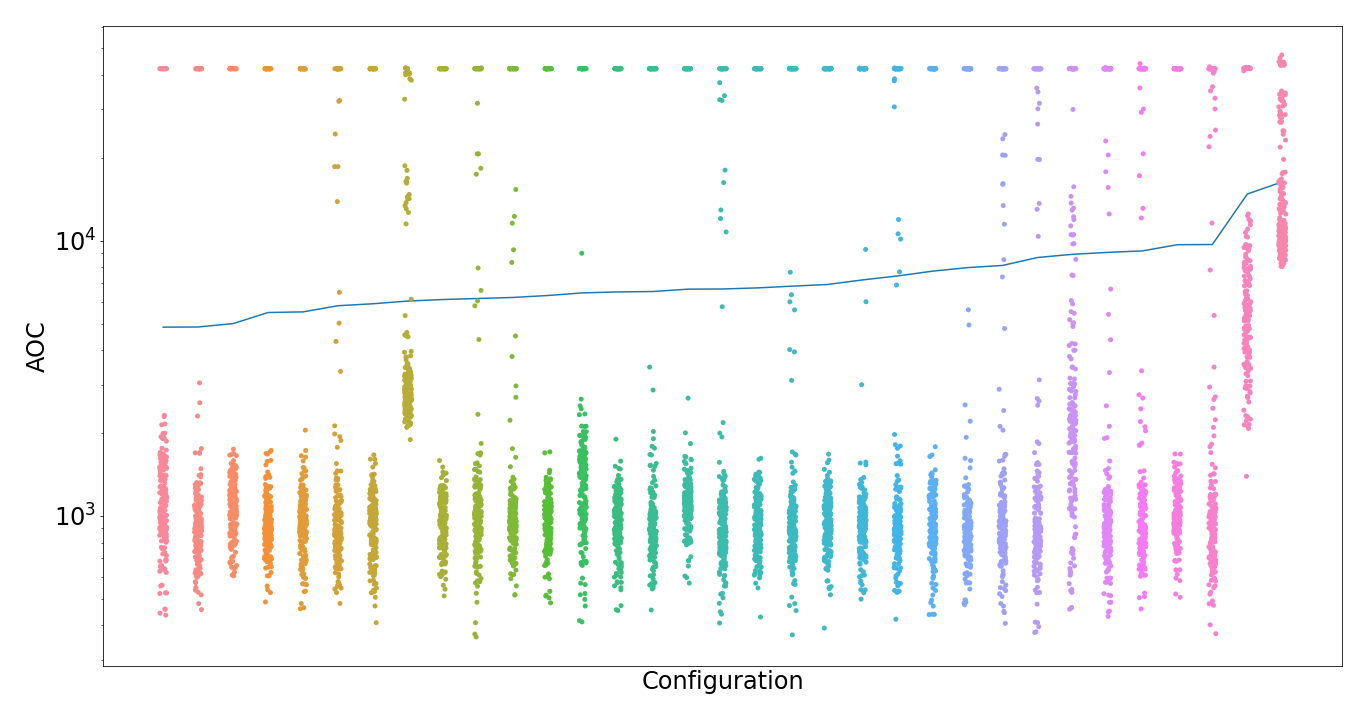}
    \caption{Distribution of  \AOC values of 200 individual runs of \hq on F8. The line indicates the mean \AOC value of each configuration, and is the basis for the sorting on the x-axis.}
    \label{fig:F8_conf_distributions}
\end{figure}

We observe large underestimation error in almost all functions. In some functions, such as F8, the underestimation error is large even for a sample size of 50.
We also notice that large underestimation errors in Figure~\ref{fig:underestimation_mean_modcma} often coincide with a large performance loss seen in Figure~\ref{fig:perf_loss_mean_modcma}. This observation can be explained by looking in more detail at the performance distribution of the used configurations on a particular function, as is done in Figure~\ref{fig:F8_conf_distributions} for F8.

We see in this figure that all configurations have a fraction of runs where the \AOC value is very large, indicating that these were very poorly performing runs. When calculating the mean value of a configuration from a limited number of samples, if none of these poor runs appears in the samples, then the mean of the configuration will be lower than its true mean, leading to the large underestimation seen in Figure~\ref{fig:underestimation_mean_modcma}. 

Additionally, since the difference in a configurations performance seen during configuration and its true mean is often larger than the difference in the means of configurations as estimated from a small number of samples, a relatively poor performing configuration can end up being chosen simply because it got `lucky', which can explain the performance losses we observed previously. 

\begin{figure*}
     \centering
     \begin{subfigure}[b]{0.32\textwidth}
         \centering
         \includegraphics[width=\textwidth, trim={5mm, 10mm, 5mm, 5mm},clip]{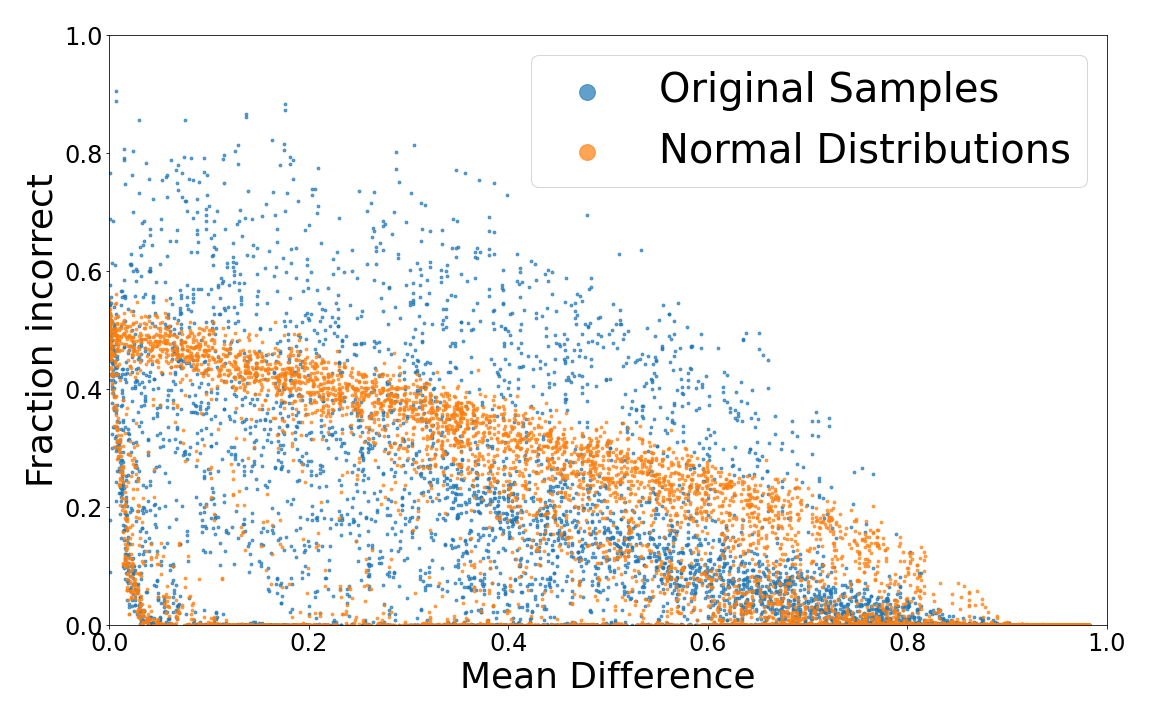}
         \caption{F9}
         \label{fig:modcma_comp_mean_samples_f9}
     \end{subfigure}
     \hfill
     \begin{subfigure}[b]{0.32\textwidth}
         \centering
         \includegraphics[width=\textwidth, trim={5mm, 10mm, 5mm, 5mm},clip]{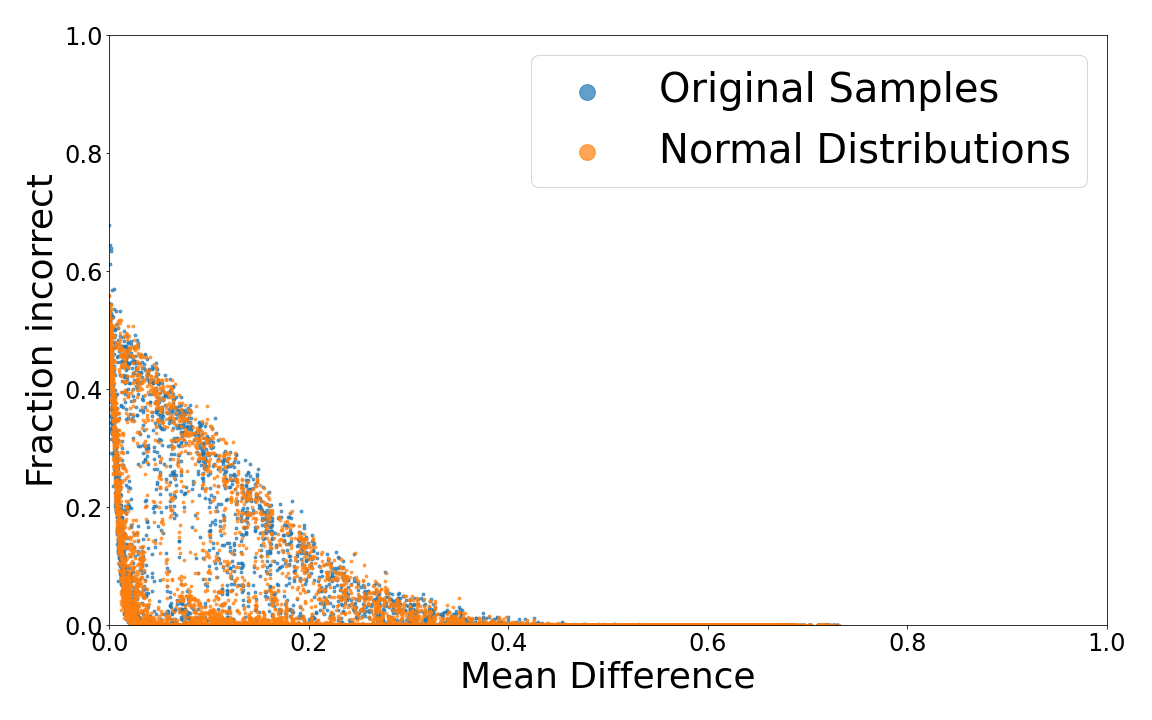}
         \caption{F15}
         \label{fig:modcma_comp_mean_samples_f15}
     \end{subfigure}
     \hfill
     \begin{subfigure}[b]{0.32\textwidth}
         \centering
         \includegraphics[width=\textwidth, trim={5mm, 10mm, 5mm, 5mm},clip]{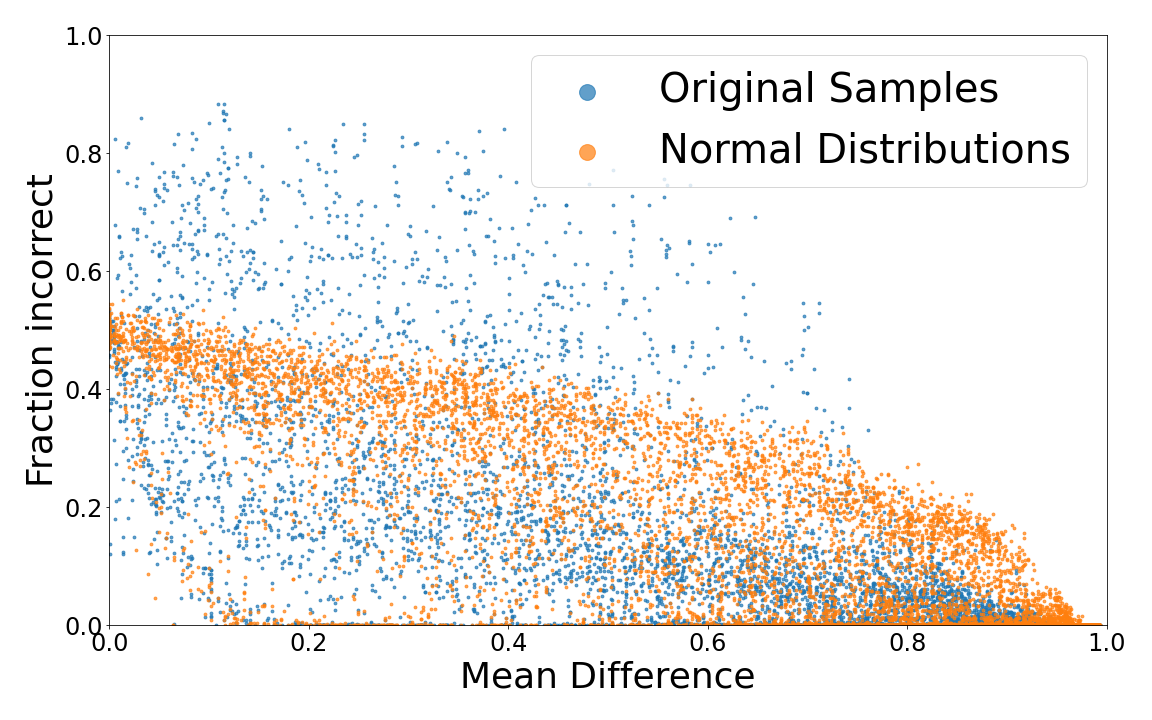}
         \caption{F22}
         \label{fig:modcma_comp_mean_samples_f22}
     \end{subfigure}
     \caption{
     Fraction of incorrect decisions when using the sample mean to compare pairs of modCMA configurations. Each subplot contains 10\,000 points. Each point compares two configurations selected uniformly at random from the available configurations. The x-axis indicates the normalized difference between their true means (based on the 200 \AOC values per configuration). The y-axis indicates the fraction of incorrect decisions based on 500 independent samplings of 15 \AUC values for each of the two selected configuration. \emph{Original samples} refers to sampling with replacement from the 200 \AOC values available, while \emph{Normal distributions} refers to sampling values from a normal distribution with the same mean and standard deviation as the 200 values of the corresponding configuration. }
        \label{fig:modcma_comp_mean_samples}
\end{figure*}

Another common way in which the mean is used in benchmarking is in the basic pairwise comparison scenario, where two algorithms are directly compared to each other. To investigate this scenario, we simulate pairwise comparisons based on a limited sample size, and correlate the decisions made by the pairwise comparison to the difference in true means between the selected configurations. To achieve this simulation, we use the full set of modCMA configurations generated during an irace run, which totals over 200 configurations on each function. From this set of configurations,  we take 10\,000 pairs, drawn uniformly at random, to perform the pairwise comparison. Then, for each pair of configurations, we  sample a number of \AOC values from the 200 values available, calculate the sample means and compare them to decide which configuration is the best. The comparison is \emph{correct} if it gives the same conclusion as comparing the true means. We repeat the sampling and comparison step 500 times to calculate the fraction of times that the comparison is correct.
The results of this experiment on F9, F15, and F22, with sample size 15, are displayed in Figure~\ref{fig:modcma_comp_mean_samples}. 

We observe in this figure that, as expected, the fraction of incorrect decisions decreases when the difference in true means increases. However, the decrease is much faster for  F15 than for F9 or F22. There are also notable differences when comparing the fraction of incorrect decisions generated by sampling with replacement from the 200 \AOC values available (Original samples) versus sampling values from the normal distribution that has the same mean and standard deviation as those 200 values. These distributions are almost identical for F15 but different for F9 and F22, which suggests that the fraction of incorrect decisions made by comparing means for F9 and F22 is impacted by the non-normality of the samples distribution.

\begin{figure}[tbh]
    \centering
    \includegraphics[width=0.48\textwidth, trim={5mm, 10mm, 5mm, 5mm},clip]{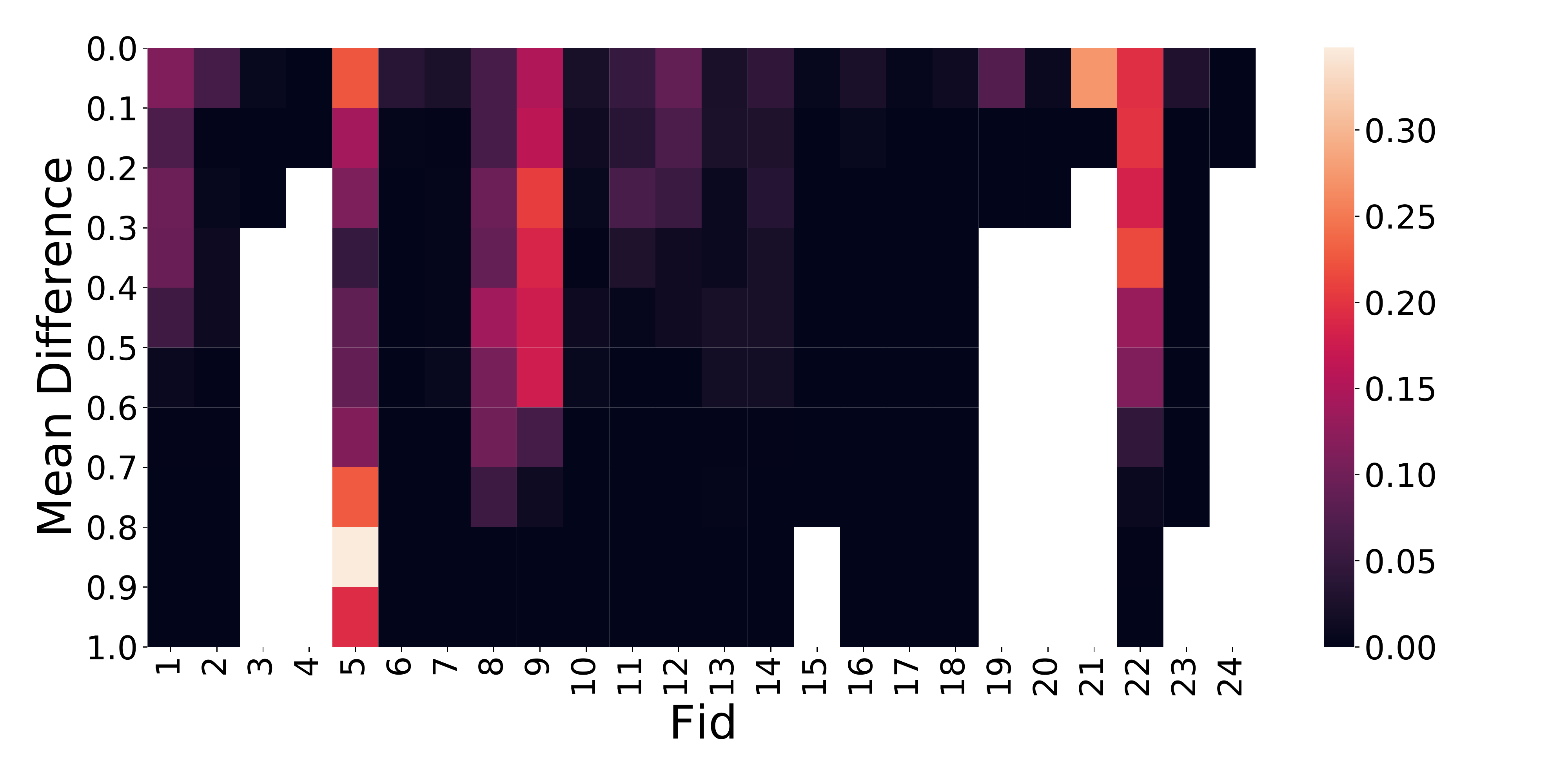}
    \caption{Fraction of configuration-pairs where the t-test gives an incorrect conclusion in more than $\alpha=0.05$ of cases, on each of the 24 BBOB functions when considering 10\,000 random pairs of modCMA configurations and a sample size of 15.}\vspace{-10pt}
    \label{fig:frac_mistakes_ttest}
\end{figure}

\begin{figure*}
     \centering
     \begin{subfigure}[b]{0.32\textwidth}
         \centering
         \includegraphics[width=\textwidth,trim=10mm 5mm 15mm 5mm, clip]{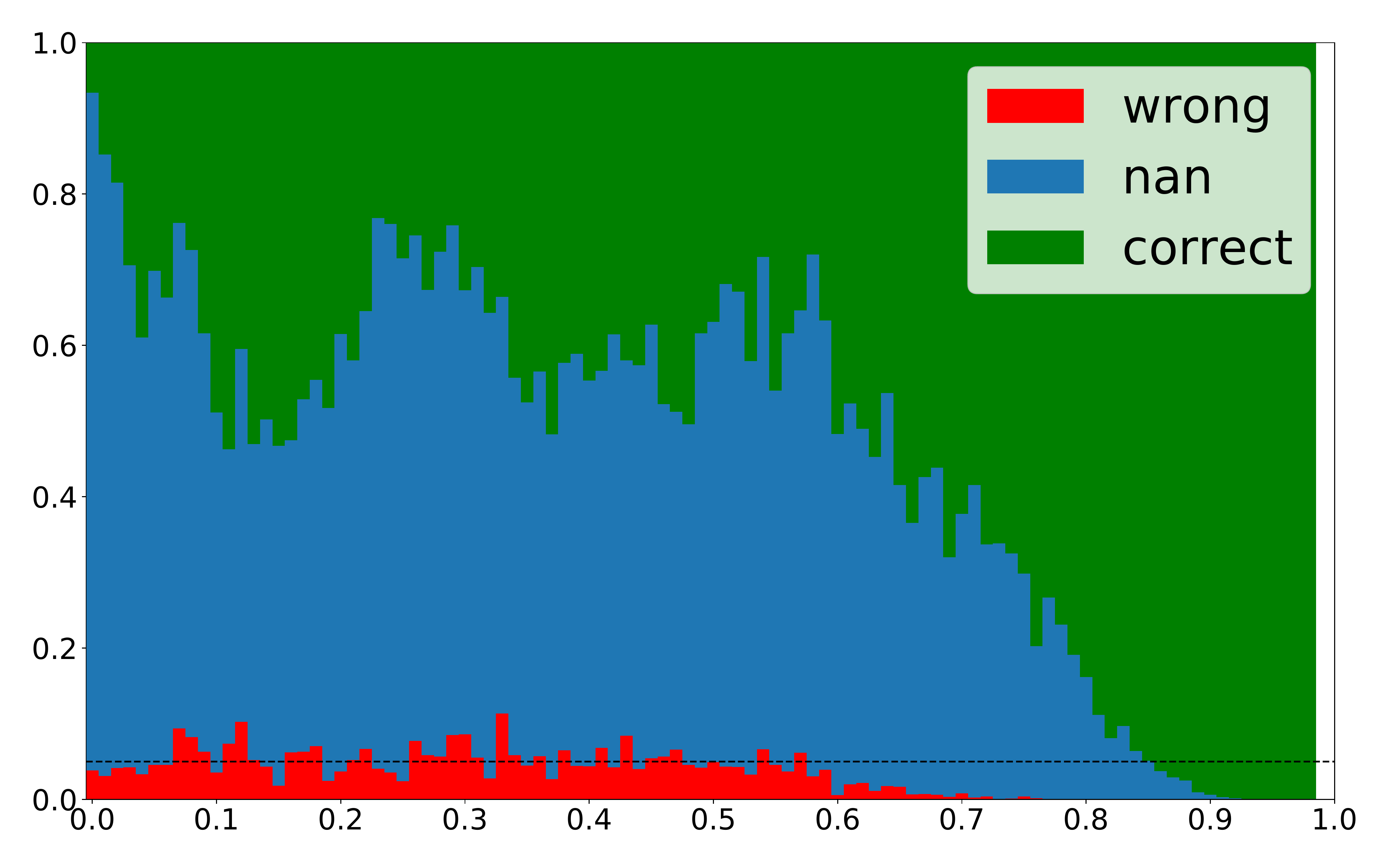}
         \caption{t-test}
         \label{fig:F9_bins_ttest}
     \end{subfigure}
     \hfill
     \begin{subfigure}[b]{0.32\textwidth}
         \centering
         \includegraphics[width=\textwidth,trim=10mm 5mm 15mm 5mm, clip]{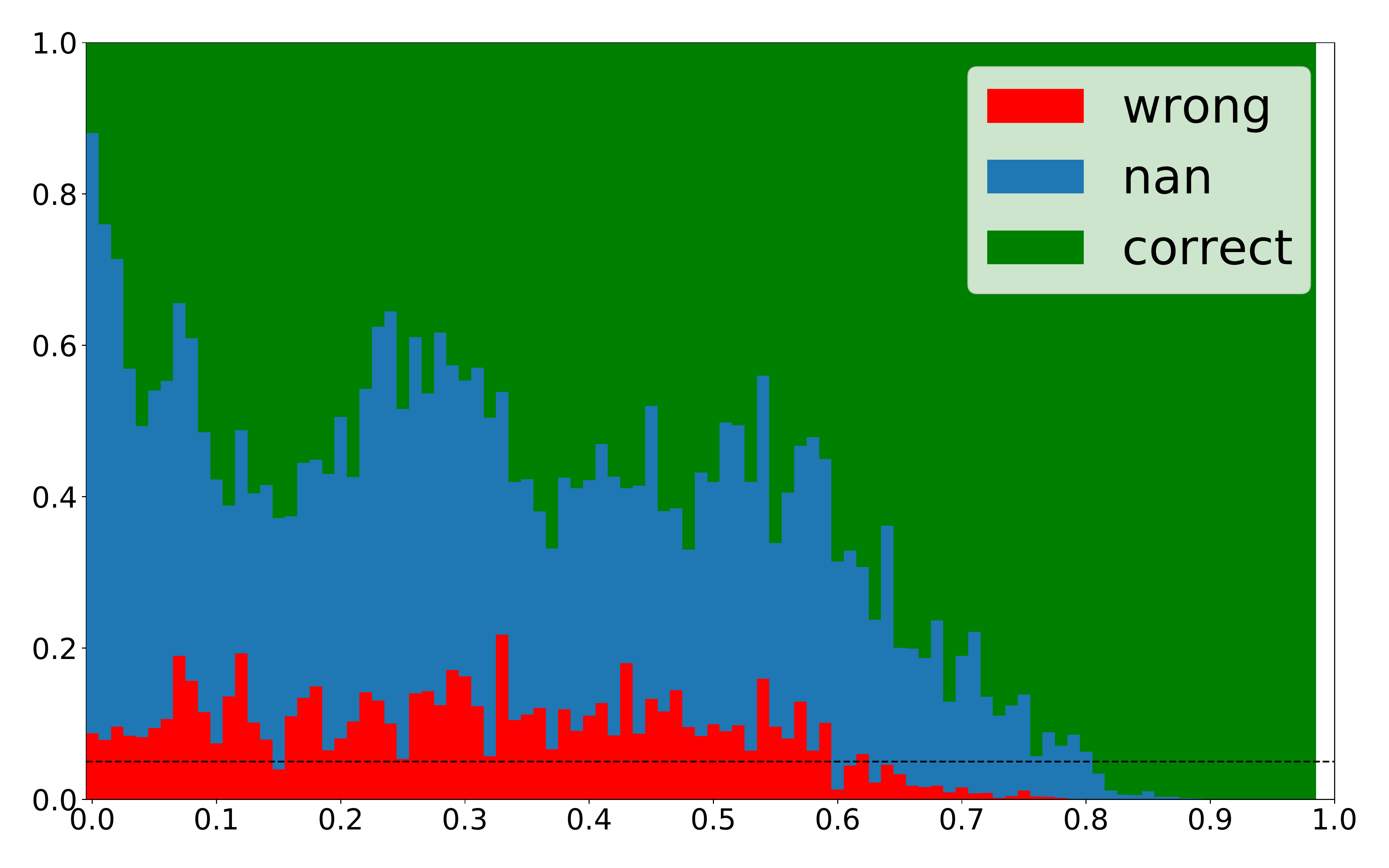}
         \caption{Wilcoxon ranked-sum test}
         \label{fig:F9_bins_wtest}
     \end{subfigure}
     \hfill
     \begin{subfigure}[b]{0.32\textwidth}
         \centering
         \includegraphics[width=\textwidth,trim=10mm 5mm 15mm 5mm, clip]{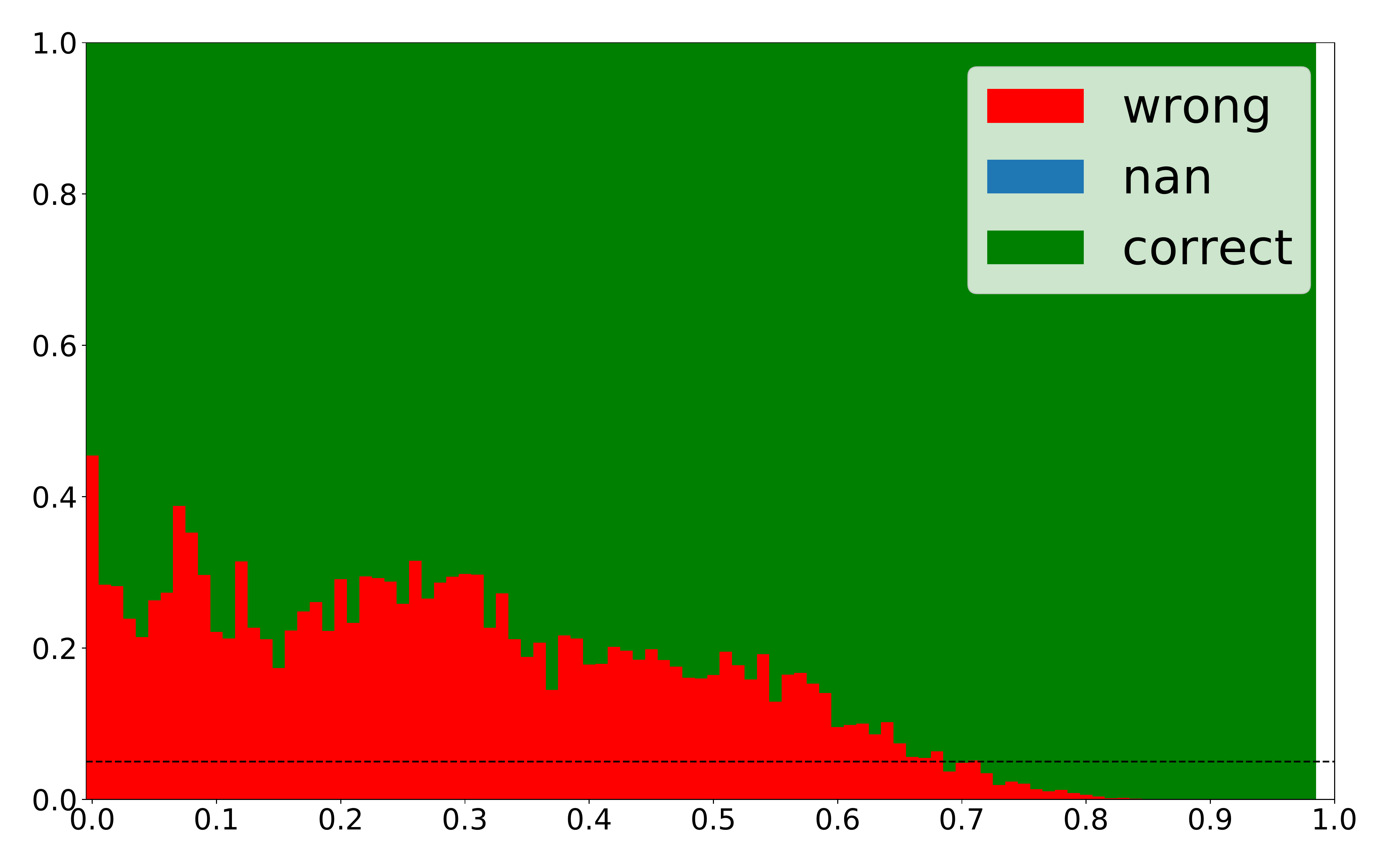}
         \caption{Comparison means}
         \label{fig:F9_bins_mtest}
     \end{subfigure}
        \caption{Correctness of decisions made in pairwise comparisons between modular CMA-ES configurations on F9, using different  procedures. The x-axis shows the relative difference in true mean  between the selected configurations. The y-axis shows the fraction of comparisons, out of 500 repetitions, that the decision was correct, incorrect or inconclusive (nan) when comparing configurations with this difference. Each repetition samples 15 values out of the 200 available for each configuration compared. These figures are available for all 24 functions and multiple sample sizes in our figshare repository~\cite{figshare_figures}.}
        \label{fig:F9_3tests_bins}
\end{figure*}

\section{Statistical testing}\label{sec:stat_test}

When considering pairwise comparisons between algorithms, we often use statistical tests to determine if one algorithm outperforms the other. Two of the most common tests are the t-test and the non-parametric Wilcoxon rank-sum test. 

To more closely analyze these two testing procedures, we re-sample with replacement, for sample size 15, from the set of 200 \AOC values of the 33 \hq.
Then, we apply a one-sided $t$-test to the samples of size 15 and measure the fraction of pairs in which the test was ``correct'', ``incorrect'' or ``inconclusive''. We consider here that the test is ``incorrect'' when, for a pair of algorithms $A$ and $B$, the null hypothesis that $A$ has a lower mean than $B$ is rejected but the mean of $A$ is indeed lower than the mean of $B$ based on the 200 values. When neither of the two one-sided null hypotheses ($A$ has lower mean than $B$ nor $B$ has lower mean than $A$) are rejected, the test is considered ``inconclusive''.

In Figure~\ref{fig:frac_mistakes_ttest}, we show the fraction of configuration pairs where the amount of incorrect tests exceeds the used level of statistical significance ($\alpha=0.05$). From this figure we see that, while the t-test seems to work well enough for most functions, it is not ideal on all functions, which seemingly indicates that the normality assumption is not met. 

We zoom in on function F9 in Figure~\ref{fig:F9_3tests_bins}, and look at the difference between making decisions based on means, t-test and Wilcoxon rank-sum test. We note that both statistical tests show an error rate that is larger than $\alpha$ for pairs of configurations with a difference in means up to $60\%$. We also note that even though the t-test is less frequently incorrect, it is also more frequently inconclusive compared to the Wilcoxon rank-sum test, even for configuration whose means differ significantly. 

Inconclusiveness is not a factor when comparing based on means, but that comes with the cost of making more incorrect decisions as well. While the number of incorrect decisions decreases when adding more samples, the overall observations for the three comparison procedures remain  similar~\cite{figshare_figures}.

\begin{figure*}[tb]
     \centering
     \begin{subfigure}[b]{0.32\textwidth}
         \centering
         \includegraphics[width=\textwidth, trim={5mm, 10mm, 5mm, 5mm},clip]{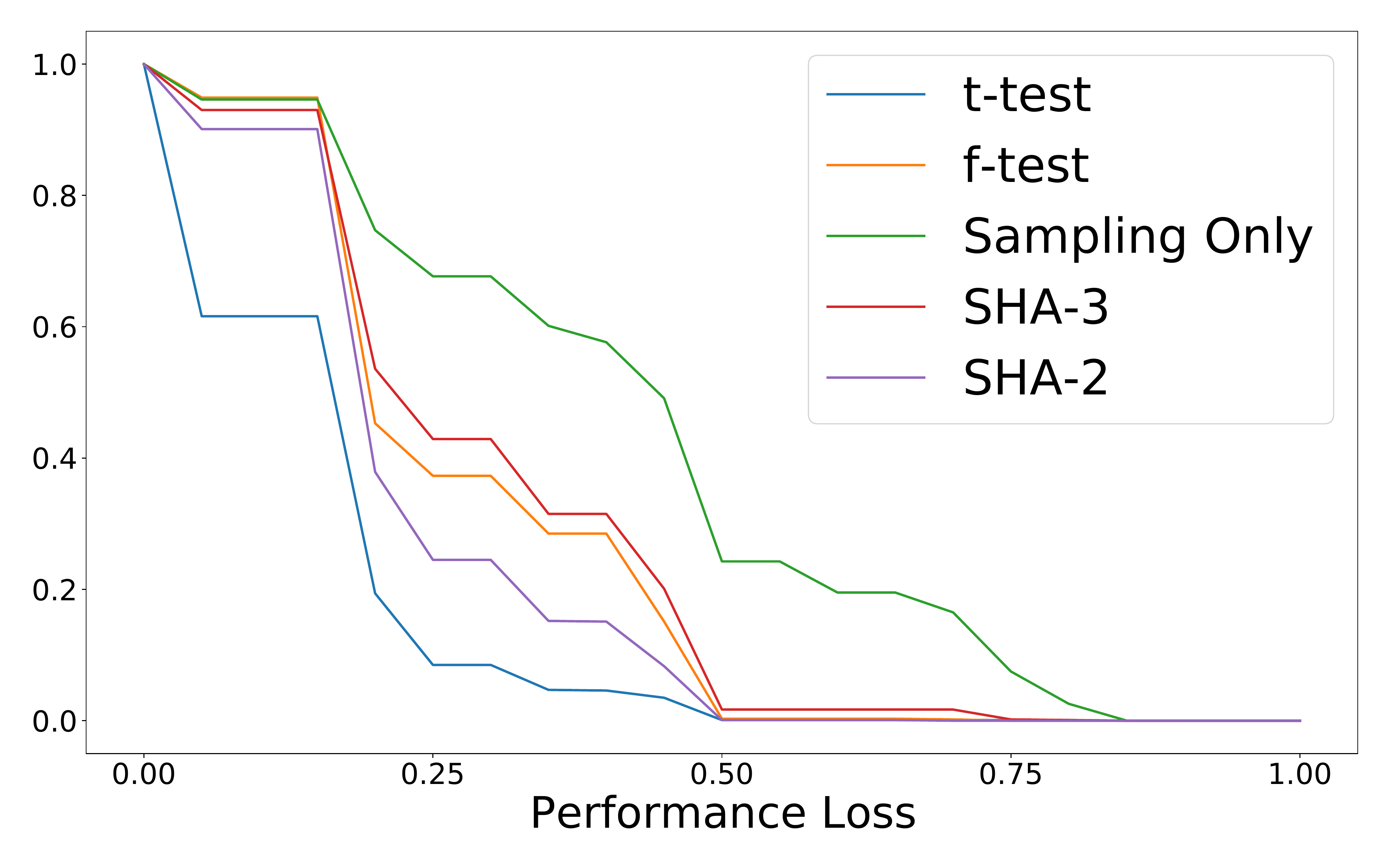}
         \caption{F9}
         \label{fig:perf_loss_cdf_F9}
     \end{subfigure}
     \hfill
     \begin{subfigure}[b]{0.32\textwidth}
         \centering
         \includegraphics[width=\textwidth, trim={5mm, 10mm, 5mm, 5mm},clip]{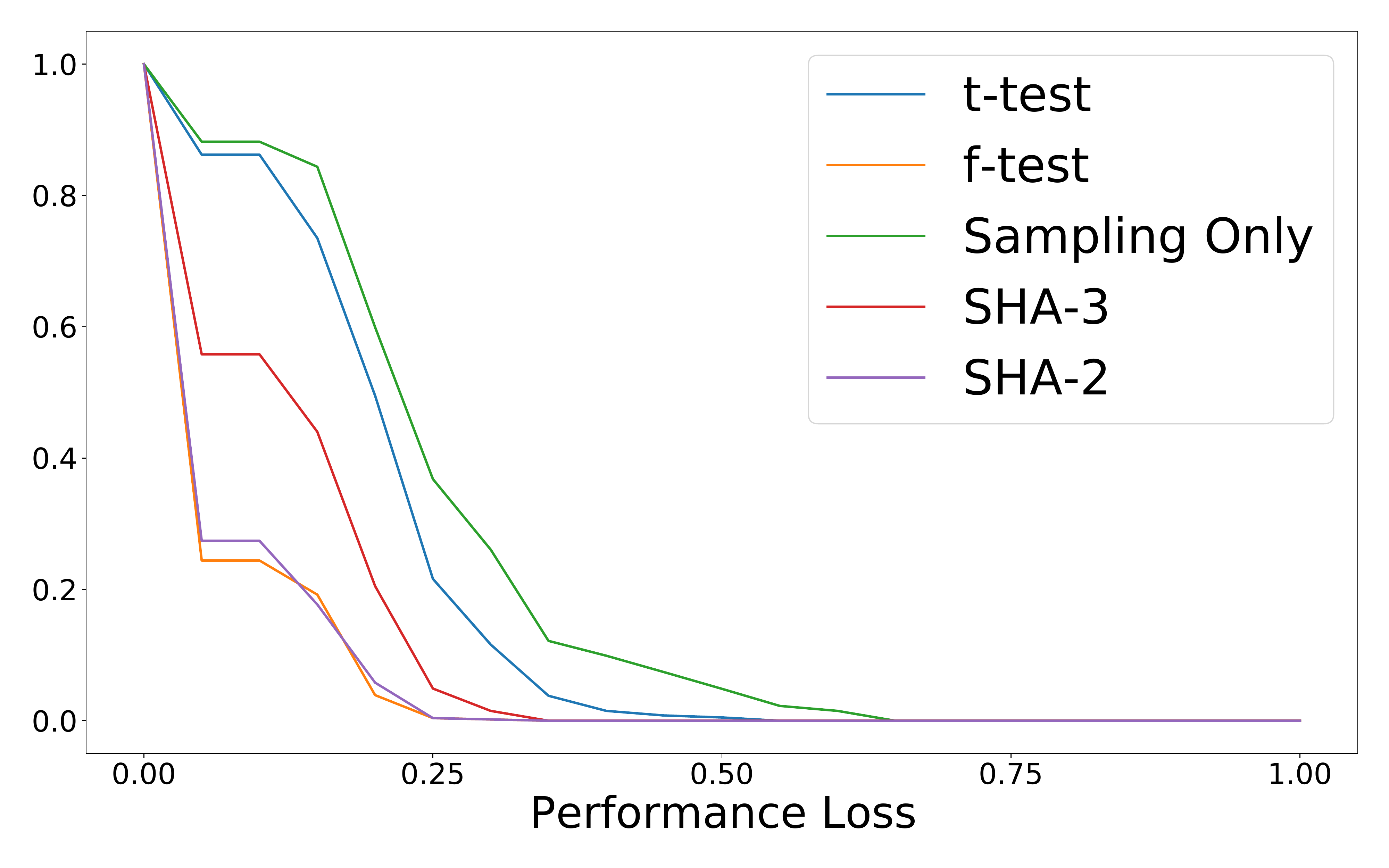}
         \caption{F15}
         \label{fig:perf_loss_cdf_F15}
     \end{subfigure}
     \hfill
     \begin{subfigure}[b]{0.32\textwidth}
         \centering
         \includegraphics[width=\textwidth, trim={5mm, 10mm, 5mm, 5mm},clip]{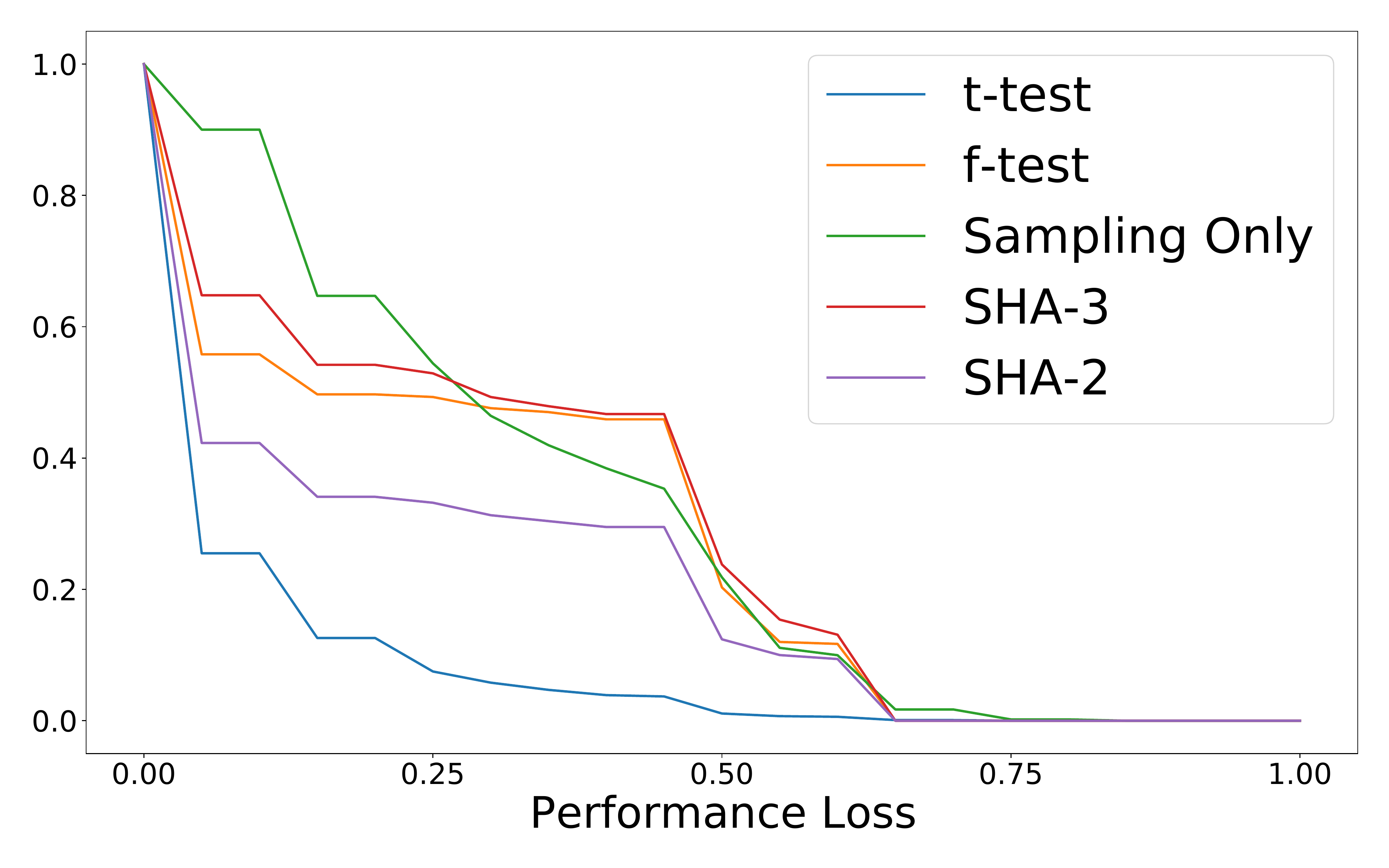}
         \caption{F22}
         \label{fig:perf_loss_cdf_F22}
     \end{subfigure}
        \caption{Cumulative performance loss of 5 variants of the racing procedure using $\textit{FirstTest}=2$: t-test, Friedman-test, sampling and selecting based on mean, and successive halving with reduction factors 2 and 3. }
        \label{fig:perf_loss_cdf}
\end{figure*}

\begin{figure*}[tb]
     \centering
     \begin{subfigure}[b]{0.32\textwidth}
         \centering
         \includegraphics[width=\textwidth, trim={5mm, 10mm, 5mm, 5mm},clip]{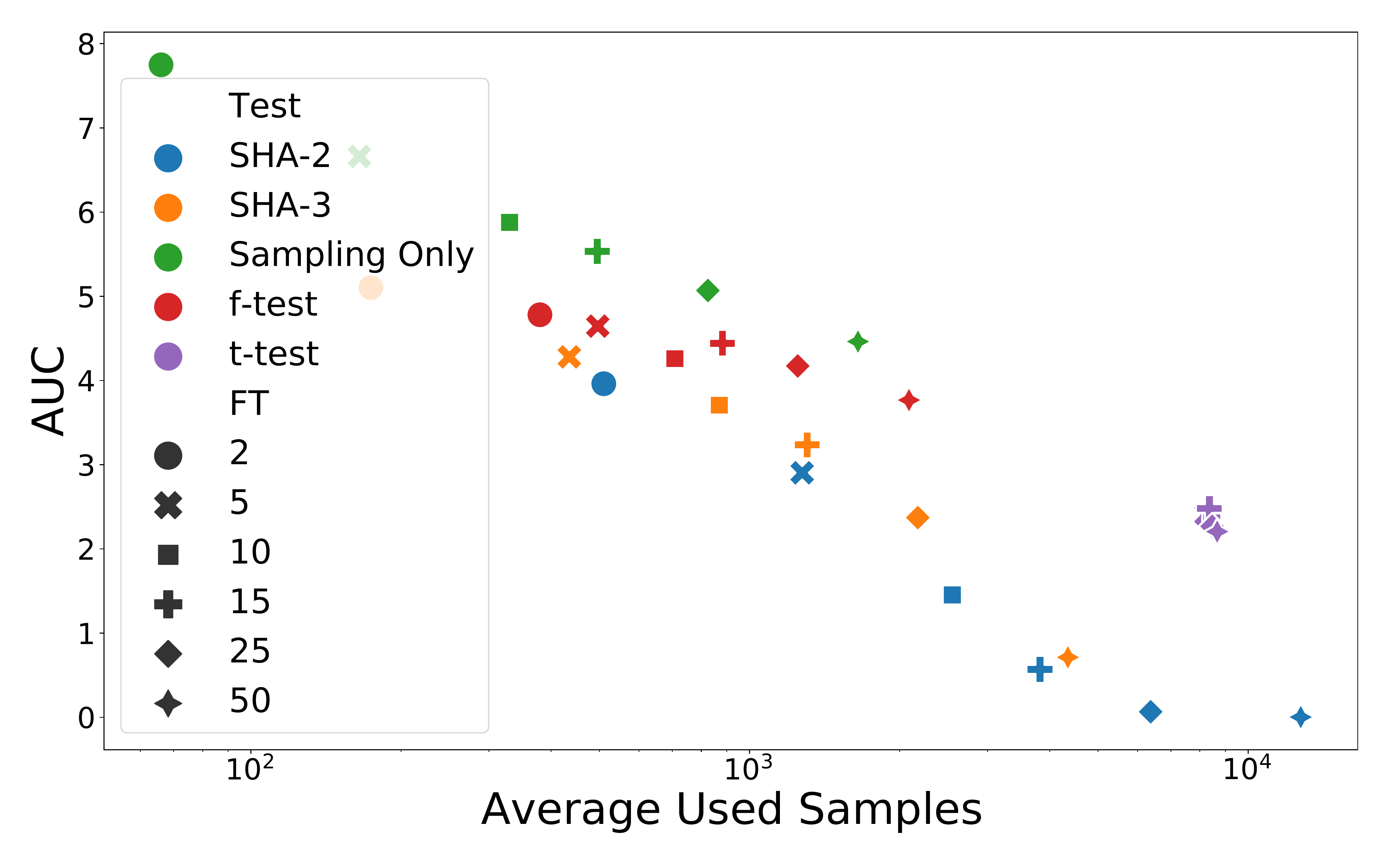}
         \caption{F9}
         \label{fig:AUC_perf_loss_F9}
     \end{subfigure}
     \hfill
     \begin{subfigure}[b]{0.32\textwidth}
         \centering
         \includegraphics[width=\textwidth, trim={5mm, 10mm, 5mm, 5mm},clip]{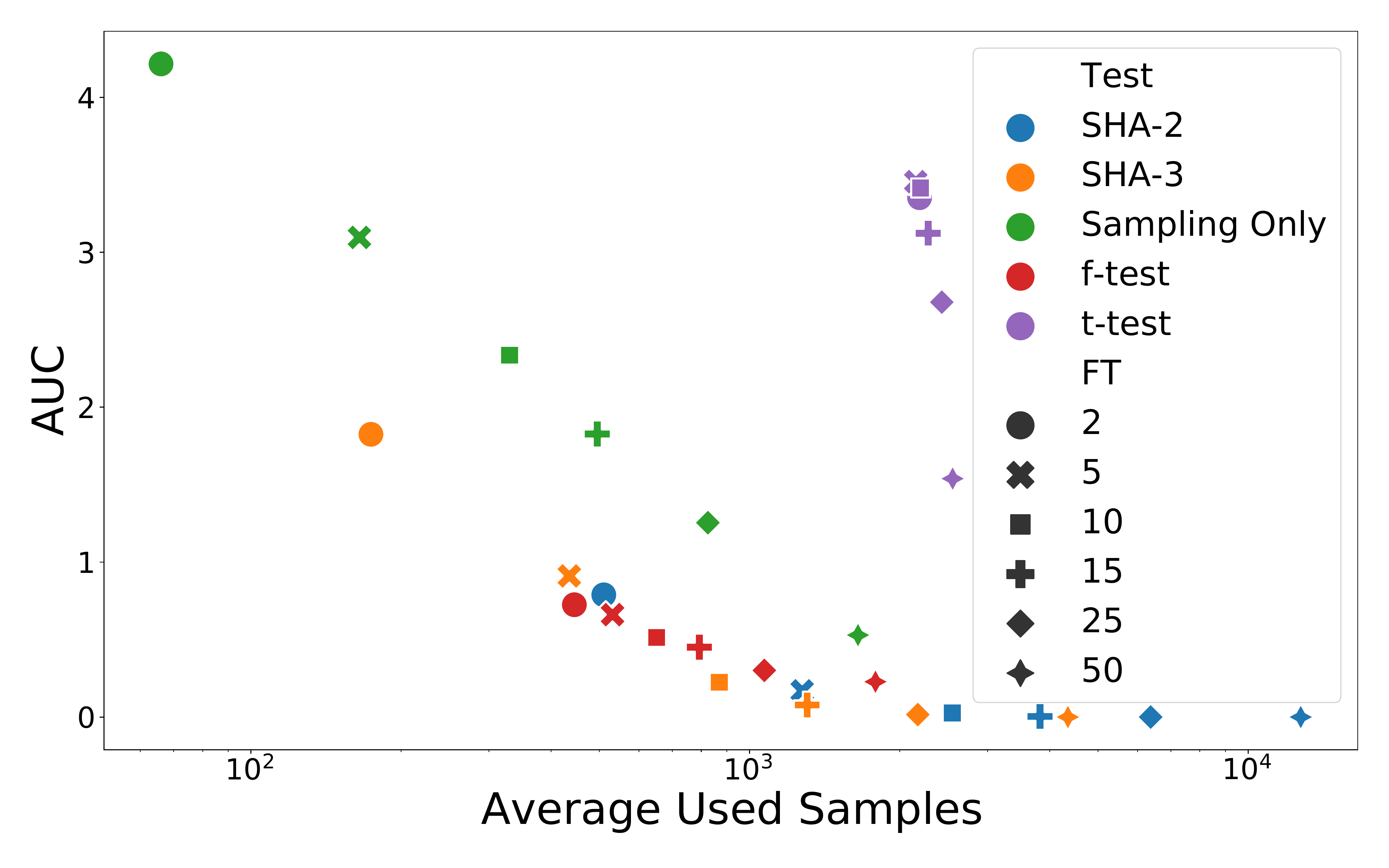}
         \caption{F15}
         \label{fig:AUC_perf_loss_F15}
     \end{subfigure}
     \hfill
     \begin{subfigure}[b]{0.32\textwidth}
         \centering
         \includegraphics[width=\textwidth, trim={5mm, 10mm, 5mm, 5mm},clip]{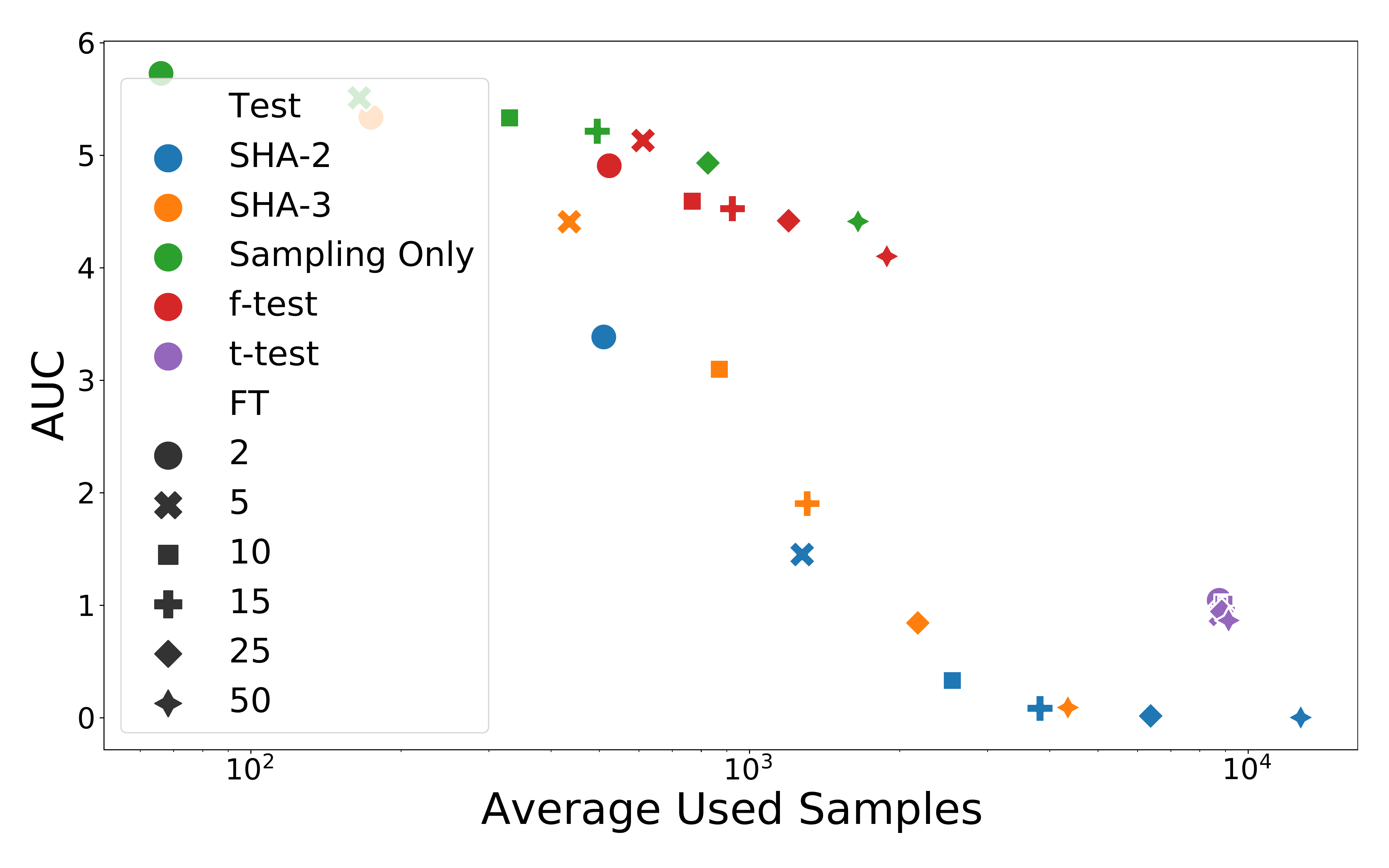}
         \caption{F22}
         \label{fig:AUC_perf_loss_F22}
     \end{subfigure}
        \caption{Comparison of AUC value of Cumulated performance loss (Figure~\ref{fig:perf_loss_cdf}), relative to the average amount of samples used by each process.}
        \label{fig:AUC_perf_loss}
\end{figure*}

\section{Racing}\label{sec:racing}
To investigate the impact of performance variability on algorithm configuration, we focus on the racing procedures used by irace, which we simulate using the \hq from Section~\ref{sec:modcma}. In particular, we consider two variants of the racing procedure~\cite{MarMoo1997air}using either the t-test or the Friedman-test. In addition to these racing variants, we also consider two variants of Successive HAlving (SHA)~\cite{KarKorSom2013} with reduction factors 2 and 3, respectively. 
For the races using statistical tests, we loosen the total budget restriction, which is usually used as stopping criteria~\cite{LopDubPerStuBir2016irace}, (e.g., in irace) to 10\,000 total samples, which means we continue the race until 5 or fewer configurations remain, or until we exceed 10\,000 sampled runs (`target runs' in irace terminology). We simulate this race 1000 times for each function and several values of \textit{FirstTest}, and show the resulting performance loss for F9, F15, and F22 in Figure~\ref{fig:perf_loss_cdf}. In this figure, performance loss is defined as the difference in true mean of the best elite (configuration with the best sampled mean during the race) against the best configuration which was present in the race.

The cumulative performance loss is compared for both the Friedman-test and t-test variants of the racing procedure, as well as a naive \emph{sampling-only} approach that selects based on means after \textit{FirstTest} samples have been collected for each configuration.  

When comparing the different approaches, we note that there is not a clear winner across all functions and values of \textit{FirstTest}. Interestingly, for some of the functions where Figure~\ref{fig:irace_perf_loss_all} shows the largest performance losses of irace elites, the races using the Friedman test seem to perform relatively poorly. This might indicate that for these functions, we could regain some of the lost performance, if it can be detected during the algorithm configuration that a different testing strategy would be required.

From Figure~\ref{fig:perf_loss_cdf}, we can clearly see that any any variant of racing or SHA is much more reliable than  the \emph{sampling-only} approach. However, racing uses more total samples, since it add runs when needed, while the sampling-only approach uses a fixed number of samples. The SHA method uses a fixed number of samples as well, but this number is significantly larger than the sampling-only approach, and depends on the reduction factor used. 

In order to account for the differences in total budget, we 

summarize cumulative performance loss curves, such as those in Figure~\ref{fig:perf_loss_cdf}, using their corresponding AUC values, and plot  these AUC values against the total samples used in Figure~\ref{fig:AUC_perf_loss}.

Here, we see an explanation for the great performance of the t-test: it uses significantly more samples for the same \textit{FirsTest} value than any of the other methods. This can happen when the test can not make any conclusive decision between the configurations, and thus fails to reject enough configurations to reach the $5$ elites, using up the full budget of 10\,000 evaluations in the process. This matches our findings from Figure~\ref{fig:F9_bins_ttest}, where we could see that the pairwise t-test often does not give any decision, even when the difference in true means  between configuration is relatively large. 

\section{Conclusions and Future Work}\label{sec:concl}

In this work, we have highlighted that commonly used performance metrics can have various non-normal distributions with large amounts of variance. While this is inherent to the field of iterative optimization heuristics, the impact of this stochasiticity is often overlooked in empirical work, where rankings between algorithms are made purely based on aggregations of limited runs, which can lead to incorrect decisions between algorithms. 

While using statistical test largely alleviates these problems, they are not a silver bullet. Since data is often shared to compare new algorithms to the state-of-the-art, continuous re-use of the same few data points has the potential to lead to bias, especially when the underlying distribution has a high variance. A similar effect can be clearly seen in algorithm configuration, where choices between similarly performing algorithms are made, and even if each individual choice is valid, the overall result is likely to significantly underestimate the true performance of the chosen configuration. This can lead to a selection of sub-optimal configurations, which can be considerably worse than others that participated in the same race.

In an ideal case, each time a comparison is done, the used samples from both algorithms should be newly generated. However, this is obviously infeasible, as computation time is a limited resource. As such, making a larger set of samples available would be a more practical solution. The recommended number of samples would differ on a per-function basis, as some functions inherently cause algorithms run on them to have a higher variance of performance. Power-analysis studies~\cite{ellis2010essential,CamWan2020sample} based on algorithms for which performance data is already available might help us find better defaults, but since we can not know what kind of distributions the collected samples will be compared against in future, this might not fully solve the problem. 

In addition, a more robust statistical analysis of the commonly used performance measures would be highly beneficial to gain more insight into the reasons for the observed errors. To aid with reproducibility, moving from standard hypothesis testing to the safe variant~\cite{grunwald2020safe}, where samples can be added continuously, would allow us to add more samples when this is deemed necessary. If this is combined with better guidelines for code availability and standards for data sharing (including formatting guidelines to ease interoperability), it will allow any researcher to gather new samples from existing algorithms to expand algorithm comparisons where needed, without hurting the statistical rigour of the comparison. 

\smallskip{}
\textbf{Reproducibility.}\label{sec:repr} 
To ensure that the work shown in this paper is reproducible~\cite{LopBraPaq2021telo}, all data and code used is made available on Zenodo~\cite{zenodo}. This includes figure generation code for figures that have not been included here because of the limited space available. For ease of viewing, these additional figures have also been uploaded on figshare~\cite{figshare_figures}. In particular, Figure~\ref{fig:F21_Aoc_distr} for all BBOB functions and additional performance metrics, Figure~\ref{fig:F8_conf_distributions} for all functions and more combinations of configurations and Figures~\ref{fig:modcma_comp_mean_samples}, \ref{fig:F9_3tests_bins}, \ref{fig:perf_loss_cdf} and \ref{fig:AUC_perf_loss} for all functions and different sample sizes.

\begin{acks}
M.\@ L\'opez-Ib\'a\~nez is a ``Beatriz Galindo'' Senior Distinguished Researcher (BEAGAL 18/00053) funded by the Spanish Ministry of Science and Innovation (MICINN). This work is supported by the Paris Ile-de-France region, via the DIM RFSI AlgoSelect project. 
\end{acks}
\bibliographystyle{ACM-Reference-Format}


\providecommand{\MaxMinAntSystem}{{$\cal MAX$--$\cal MIN$} {Ant} {System}}
  \providecommand{\rpackage}[1]{{#1}}
  \providecommand{\softwarepackage}[1]{{#1}}
  \providecommand{\proglang}[1]{{#1}}
\begin{thebibliography}{23}


\ifx \showCODEN    \undefined \def \showCODEN     #1{\unskip}     \fi
\ifx \showDOI      \undefined \def \showDOI       #1{#1}\fi
\ifx \showISBNx    \undefined \def \showISBNx     #1{\unskip}     \fi
\ifx \showISBNxiii \undefined \def \showISBNxiii  #1{\unskip}     \fi
\ifx \showISSN     \undefined \def \showISSN      #1{\unskip}     \fi
\ifx \showLCCN     \undefined \def \showLCCN      #1{\unskip}     \fi
\ifx \shownote     \undefined \def \shownote      #1{#1}          \fi
\ifx \showarticletitle \undefined \def \showarticletitle #1{#1}   \fi
\ifx \showURL      \undefined \def \showURL       {\relax}        \fi
\providecommand\bibfield[2]{#2}
\providecommand\bibinfo[2]{#2}
\providecommand\natexlab[1]{#1}
\providecommand\showeprint[2][]{arXiv:#2}

\bibitem[\protect\citeauthoryear{Bartz-Beielstein, Doerr, van~den Berg, Bossek,
  Chandrasekaran, Eftimov, Fischbach, Kerschke, {La Cava},
  L{\'o}pez-Ib{\'a}{\~n}ez, Malan, Moore, Naujoks, Orzechowski, Volz, Wagner,
  and Weise}{Bartz-Beielstein et~al\mbox{.}}{2020}]%
        {BarDoeBer2020benchmarking}
\bibfield{author}{\bibinfo{person}{Thomas Bartz-Beielstein},
  \bibinfo{person}{Carola Doerr}, \bibinfo{person}{Daan van~den Berg},
  \bibinfo{person}{Jakob Bossek}, \bibinfo{person}{Sowmya Chandrasekaran},
  \bibinfo{person}{Tome Eftimov}, \bibinfo{person}{Andreas Fischbach},
  \bibinfo{person}{Pascal Kerschke}, \bibinfo{person}{William {La Cava}},
  \bibinfo{person}{Manuel L{\'o}pez-Ib{\'a}{\~n}ez},
  \bibinfo{person}{Katherine~M. Malan}, \bibinfo{person}{Jason~H. Moore},
  \bibinfo{person}{Boris Naujoks}, \bibinfo{person}{Patryk Orzechowski},
  \bibinfo{person}{Vanessa Volz}, \bibinfo{person}{Markus Wagner}, {and}
  \bibinfo{person}{Thomas Weise}.} \bibinfo{year}{2020}\natexlab{}.
\newblock \showarticletitle{Benchmarking in Optimization: Best Practice and
  Open Issues}.
\newblock \bibinfo{journal}{\emph{Arxiv preprint arXiv:2007.03488 [cs.NE]}}
  (\bibinfo{year}{2020}).
\newblock
\urldef\tempurl%
\url{https://arxiv.org/abs/2007.03488}
\showURL{%
\tempurl}


\bibitem[\protect\citeauthoryear{Campelo and Wanner}{Campelo and
  Wanner}{2020}]%
        {CamWan2020sample}
\bibfield{author}{\bibinfo{person}{Felipe Campelo} {and}
  \bibinfo{person}{Elizabeth~F. Wanner}.} \bibinfo{year}{2020}\natexlab{}.
\newblock \showarticletitle{Sample size calculations for the experimental
  comparison of multiple algorithms on multiple problem instances}.
\newblock \bibinfo{journal}{\emph{Journal of Heuristics}} \bibinfo{volume}{26},
  \bibinfo{number}{6} (\bibinfo{year}{2020}), \bibinfo{pages}{851--883}.
\newblock
\urldef\tempurl%
\url{https://doi.org/10.1007/s10732-020-09454-w}
\showDOI{\tempurl}


\bibitem[\protect\citeauthoryear{de~Nobel, Vermetten, Wang, Doerr, and
  B{\"a}ck}{de~Nobel et~al\mbox{.}}{2021a}]%
        {NobVerWan2021gecco-supp}
\bibfield{author}{\bibinfo{person}{Jacob de Nobel}, \bibinfo{person}{Diederick
  Vermetten}, \bibinfo{person}{Hao Wang}, \bibinfo{person}{Carola Doerr}, {and}
  \bibinfo{person}{Thomas B{\"a}ck}.} \bibinfo{year}{2021}\natexlab{a}.
\newblock \bibinfo{title}{Data and Code from Tuning as a means of assessing the
  benefits of new ideas in interplay with existing algorithmic modules}.
\newblock
\newblock
\urldef\tempurl%
\url{https://doi.org/10.5281/zenodo.4524959}
\showDOI{\tempurl}


\bibitem[\protect\citeauthoryear{de~Nobel, Vermetten, Wang, Doerr, and
  B{\"a}ck}{de~Nobel et~al\mbox{.}}{2021b}]%
        {NobVerWan2021gecco}
\bibfield{author}{\bibinfo{person}{Jacob de Nobel}, \bibinfo{person}{Diederick
  Vermetten}, \bibinfo{person}{Hao Wang}, \bibinfo{person}{Carola Doerr}, {and}
  \bibinfo{person}{Thomas B{\"a}ck}.} \bibinfo{year}{2021}\natexlab{b}.
\newblock \showarticletitle{Tuning as a means of assessing the benefits of new
  ideas in interplay with existing algorithmic modules}.
\newblock In \bibinfo{booktitle}{\emph{Proceedings of the Genetic and
  Evolutionary Computation Conference Companion, GECCO Companion 2021}},
  \bibfield{editor}{\bibinfo{person}{Francisco Chicano} {and}
  \bibinfo{person}{Krzysztof Krawiec}} (Eds.). \bibinfo{publisher}{ACM Press},
  \bibinfo{address}{New York, NY}, \bibinfo{pages}{1375--1384}.
\newblock
\urldef\tempurl%
\url{https://doi.org/10.1145/3449726.3463167}
\showDOI{\tempurl}


\bibitem[\protect\citeauthoryear{Doerr, Wang, Ye, van Rijn, and B{\"a}ck}{Doerr
  et~al\mbox{.}}{2018}]%
        {IOHprofiler}
\bibfield{author}{\bibinfo{person}{Carola Doerr}, \bibinfo{person}{Hao Wang},
  \bibinfo{person}{Furong Ye}, \bibinfo{person}{Sander van Rijn}, {and}
  \bibinfo{person}{Thomas B{\"a}ck}.} \bibinfo{year}{2018}\natexlab{}.
\newblock \showarticletitle{{IOHprofiler}: A Benchmarking and Profiling Tool
  for Iterative Optimization Heuristics}.
\newblock \bibinfo{journal}{\emph{Arxiv preprint arXiv:1806.07555}}
  (\bibinfo{date}{Oct.} \bibinfo{year}{2018}).
\newblock
\urldef\tempurl%
\url{https://doi.org/10.48550/arXiv.1810.05281}
\showDOI{\tempurl}


\bibitem[\protect\citeauthoryear{Ellis}{Ellis}{2010}]%
        {ellis2010essential}
\bibfield{author}{\bibinfo{person}{Paul~D Ellis}.}
  \bibinfo{year}{2010}\natexlab{}.
\newblock \bibinfo{booktitle}{\emph{The essential guide to effect sizes:
  Statistical power, meta-analysis, and the interpretation of research
  results}}.
\newblock \bibinfo{publisher}{Cambridge university press}.
\newblock


\bibitem[\protect\citeauthoryear{Gr{\"u}nwald, de~Heide, and
  Koolen}{Gr{\"u}nwald et~al\mbox{.}}{2020}]%
        {grunwald2020safe}
\bibfield{author}{\bibinfo{person}{Peter Gr{\"u}nwald}, \bibinfo{person}{Rianne
  de Heide}, {and} \bibinfo{person}{Wouter~M Koolen}.}
  \bibinfo{year}{2020}\natexlab{}.
\newblock \showarticletitle{Safe testing}. In \bibinfo{booktitle}{\emph{2020
  Information Theory and Applications Workshop (ITA)}}. IEEE,
  \bibinfo{pages}{1--54}.
\newblock


\bibitem[\protect\citeauthoryear{Hansen, Auger, Ros, Mersmann, Tu{\v s}ar, and
  Brockhoff}{Hansen et~al\mbox{.}}{2020}]%
        {HanAugMer2020coco}
\bibfield{author}{\bibinfo{person}{Nikolaus Hansen}, \bibinfo{person}{Anne
  Auger}, \bibinfo{person}{Raymond Ros}, \bibinfo{person}{Olaf Mersmann},
  \bibinfo{person}{Tea Tu{\v s}ar}, {and} \bibinfo{person}{Dimo Brockhoff}.}
  \bibinfo{year}{2020}\natexlab{}.
\newblock \showarticletitle{{COCO}: A platform for comparing continuous
  optimizers in a black-box setting}.
\newblock \bibinfo{journal}{\emph{Optimization Methods and Software}}
  \bibinfo{volume}{36}, \bibinfo{number}{1} (\bibinfo{year}{2020}),
  \bibinfo{pages}{1--31}.
\newblock
\urldef\tempurl%
\url{https://doi.org/10.1080/10556788.2020.1808977}
\showDOI{\tempurl}


\bibitem[\protect\citeauthoryear{Hansen, Finck, Ros, and Auger}{Hansen
  et~al\mbox{.}}{2009}]%
        {HanFinRosAug2009bbob}
\bibfield{author}{\bibinfo{person}{Nikolaus Hansen}, \bibinfo{person}{Steffen
  Finck}, \bibinfo{person}{Raymond Ros}, {and} \bibinfo{person}{Anne Auger}.}
  \bibinfo{year}{2009}\natexlab{}.
\newblock \bibinfo{booktitle}{\emph{Real-Parameter Black-Box Optimization
  Benchmarking 2009: Noiseless Functions Definitions}}.
\newblock \bibinfo{type}{{T}echnical {R}eport} RR-6829.
  \bibinfo{institution}{INRIA, France}.
\newblock
\newblock
\shownote{Updated February 2010}.


\bibitem[\protect\citeauthoryear{Hansen and Ostermeier}{Hansen and
  Ostermeier}{1996}]%
        {HanOst1996cma}
\bibfield{author}{\bibinfo{person}{Nikolaus Hansen} {and}
  \bibinfo{person}{Andreas Ostermeier}.} \bibinfo{year}{1996}\natexlab{}.
\newblock \showarticletitle{Adapting arbitrary normal mutation distributions in
  evolution strategies: the covariance matrix adaptation}.
\newblock In \bibinfo{booktitle}{\emph{Proceedings of the 1996 IEEE
  International Conference on Evolutionary Computation (ICEC'96)}},
  \bibfield{editor}{\bibinfo{person}{Thomas B{\"a}ck},
  \bibinfo{person}{T.~Fukuda}, {and} \bibinfo{person}{Zbigniew Michalewicz}}
  (Eds.). \bibinfo{publisher}{IEEE Press}, \bibinfo{address}{Piscataway, NJ},
  \bibinfo{pages}{312--317}.
\newblock
\urldef\tempurl%
\url{https://doi.org/10.1109/ICEC.1996.542381}
\showDOI{\tempurl}


\bibitem[\protect\citeauthoryear{Huang, Li, and Yao}{Huang
  et~al\mbox{.}}{2020}]%
        {HuaLiYao2020algconf}
\bibfield{author}{\bibinfo{person}{Changwu Huang}, \bibinfo{person}{Yuanxiang
  Li}, {and} \bibinfo{person}{Xin Yao}.} \bibinfo{year}{2020}\natexlab{}.
\newblock \showarticletitle{A Survey of Automatic Parameter Tuning Methods for
  Metaheuristics}.
\newblock \bibinfo{journal}{\emph{IEEE Transactions on Evolutionary
  Computation}} \bibinfo{volume}{24}, \bibinfo{number}{2}
  (\bibinfo{year}{2020}), \bibinfo{pages}{201--216}.
\newblock
\urldef\tempurl%
\url{https://doi.org/10.1109/TEVC.2019.2921598}
\showDOI{\tempurl}


\bibitem[\protect\citeauthoryear{Karnin, Koren, and Somekh}{Karnin
  et~al\mbox{.}}{2013}]%
        {KarKorSom2013}
\bibfield{author}{\bibinfo{person}{Zohar Karnin}, \bibinfo{person}{Tomer
  Koren}, {and} \bibinfo{person}{Oren Somekh}.}
  \bibinfo{year}{2013}\natexlab{}.
\newblock \showarticletitle{Almost optimal exploration in multi-armed bandits}.
  In \bibinfo{booktitle}{\emph{Proceedings of the 30th International Conference
  on Machine Learning, {ICML} 2013}}, \bibfield{editor}{\bibinfo{person}{Sanjoy
  Dasgupta} {and} \bibinfo{person}{David McAllester}} (Eds.),
  Vol.~\bibinfo{volume}{28}. \bibinfo{pages}{1238--1246}.
\newblock
\urldef\tempurl%
\url{http://jmlr.org/proceedings/papers/v28/}
\showURL{%
\tempurl}


\bibitem[\protect\citeauthoryear{Kerschke, Hoos, Neumann, and
  Trautmann}{Kerschke et~al\mbox{.}}{2019}]%
        {KerHooNeuTra2019}
\bibfield{author}{\bibinfo{person}{Pascal Kerschke}, \bibinfo{person}{Holger~H.
  Hoos}, \bibinfo{person}{Frank Neumann}, {and} \bibinfo{person}{Heike
  Trautmann}.} \bibinfo{year}{2019}\natexlab{}.
\newblock \showarticletitle{Automated Algorithm Selection: Survey and
  Perspectives}.
\newblock \bibinfo{journal}{\emph{Evolutionary Computation}}
  \bibinfo{volume}{27}, \bibinfo{number}{1} (\bibinfo{date}{March}
  \bibinfo{year}{2019}), \bibinfo{pages}{3--45}.
\newblock
\urldef\tempurl%
\url{https://doi.org/10.1162/evco_a_00242}
\showDOI{\tempurl}


\bibitem[\protect\citeauthoryear{Li, Jamieson, DeSalvo, Rostamizadeh, and
  Talwalkar}{Li et~al\mbox{.}}{2018}]%
        {LiJamSal2018hyperband}
\bibfield{author}{\bibinfo{person}{Lisha Li}, \bibinfo{person}{Kevin Jamieson},
  \bibinfo{person}{Giulia DeSalvo}, \bibinfo{person}{Afshin Rostamizadeh},
  {and} \bibinfo{person}{Ameet Talwalkar}.} \bibinfo{year}{2018}\natexlab{}.
\newblock \showarticletitle{Hyperband: A Novel Bandit-Based Approach to
  Hyperparameter Optimization}.
\newblock \bibinfo{journal}{\emph{Journal of Machine Learning Research}}
  \bibinfo{volume}{18}, \bibinfo{number}{185} (\bibinfo{year}{2018}),
  \bibinfo{pages}{1--52}.
\newblock


\bibitem[\protect\citeauthoryear{L{\'o}pez-Ib{\'a}{\~n}ez, Branke, and
  Paquete}{L{\'o}pez-Ib{\'a}{\~n}ez et~al\mbox{.}}{2021}]%
        {LopBraPaq2021telo}
\bibfield{author}{\bibinfo{person}{Manuel L{\'o}pez-Ib{\'a}{\~n}ez},
  \bibinfo{person}{J{\"u}rgen Branke}, {and} \bibinfo{person}{Lu{\'i}s
  Paquete}.} \bibinfo{year}{2021}\natexlab{}.
\newblock \showarticletitle{Reproducibility in Evolutionary Computation}.
\newblock \bibinfo{journal}{\emph{ACM Transactions on Evolutionary Learning and
  Optimization}} \bibinfo{volume}{1}, \bibinfo{number}{4}
  (\bibinfo{year}{2021}), \bibinfo{pages}{1--21}.
\newblock
\urldef\tempurl%
\url{https://doi.org/10.1145/3466624}
\showDOI{\tempurl}


\bibitem[\protect\citeauthoryear{L{\'o}pez-Ib{\'a}{\~n}ez, Dubois-Lacoste,
  P{\'e}rez~C{\'a}ceres, St{\"u}tzle, and Birattari}{L{\'o}pez-Ib{\'a}{\~n}ez
  et~al\mbox{.}}{2016}]%
        {LopDubPerStuBir2016irace}
\bibfield{author}{\bibinfo{person}{Manuel L{\'o}pez-Ib{\'a}{\~n}ez},
  \bibinfo{person}{J{\'e}r{\'e}mie Dubois-Lacoste}, \bibinfo{person}{Leslie
  P{\'e}rez~C{\'a}ceres}, \bibinfo{person}{Thomas St{\"u}tzle}, {and}
  \bibinfo{person}{Mauro Birattari}.} \bibinfo{year}{2016}\natexlab{}.
\newblock \showarticletitle{The {\rpackage{irace}} Package: Iterated Racing for
  Automatic Algorithm Configuration}.
\newblock \bibinfo{journal}{\emph{Operations Research Perspectives}}
  \bibinfo{volume}{3} (\bibinfo{year}{2016}), \bibinfo{pages}{43--58}.
\newblock
\urldef\tempurl%
\url{https://doi.org/10.1016/j.orp.2016.09.002}
\showDOI{\tempurl}


\bibitem[\protect\citeauthoryear{Maron and Moore}{Maron and Moore}{1997}]%
        {MarMoo1997air}
\bibfield{author}{\bibinfo{person}{O. Maron} {and} \bibinfo{person}{A.~W.
  Moore}.} \bibinfo{year}{1997}\natexlab{}.
\newblock \showarticletitle{The Racing Algorithm: {Model} Selection for Lazy
  Learners}.
\newblock \bibinfo{journal}{\emph{Artificial Intelligence Research}}
  \bibinfo{volume}{11}, \bibinfo{number}{1--5} (\bibinfo{year}{1997}),
  \bibinfo{pages}{193--225}.
\newblock


\bibitem[\protect\citeauthoryear{Rapin and Teytaud}{Rapin and Teytaud}{2018}]%
        {nevergrad}
\bibfield{author}{\bibinfo{person}{J. Rapin} {and} \bibinfo{person}{O.
  Teytaud}.} \bibinfo{year}{2018}\natexlab{}.
\newblock \bibinfo{title}{Nevergrad: A gradient-free optimization platform}.
\newblock
  \bibinfo{howpublished}{\url{https://GitHub.com/FacebookResearch/Nevergrad}}.
\newblock


\bibitem[\protect\citeauthoryear{van Rijn}{van Rijn}{2018}]%
        {modCMA}
\bibfield{author}{\bibinfo{person}{Sander van Rijn}.}
  \bibinfo{year}{2018}\natexlab{}.
\newblock \bibinfo{title}{Modular {CMA-ES} framework
  from~\cite{RijWanLeeBac2016ssci}, v0.3.0}.
\newblock \bibinfo{howpublished}{\url{https://github.com/sjvrijn/ModEA}}.
\newblock
\newblock
\shownote{Available also as {\softwarepackage{pypi}} package at
  \url{https://pypi.org/project/ModEA/0.3.0/}}.


\bibitem[\protect\citeauthoryear{van Rijn, Wang, van Leeuwen, and B{\"a}ck}{van
  Rijn et~al\mbox{.}}{2016}]%
        {RijWanLeeBac2016ssci}
\bibfield{author}{\bibinfo{person}{Sander van Rijn}, \bibinfo{person}{Hao
  Wang}, \bibinfo{person}{Matthijs van Leeuwen}, {and} \bibinfo{person}{Thomas
  B{\"a}ck}.} \bibinfo{year}{2016}\natexlab{}.
\newblock \showarticletitle{Evolving the structure of {Evolution}
  {Strategies}}. In \bibinfo{booktitle}{\emph{Computational Intelligence
  (SSCI), 2016 IEEE Symposium Series on}},
  \bibfield{editor}{\bibinfo{person}{Xuewen Chen} {and}
  \bibinfo{person}{Andreas Stafylopatis}} (Eds.). \bibinfo{pages}{1--8}.
\newblock
\urldef\tempurl%
\url{https://doi.org/10.1109/SSCI.2016.7850138}
\showDOI{\tempurl}


\bibitem[\protect\citeauthoryear{Vermetten, Wang, Doerr, and
  B{\"a}ck}{Vermetten et~al\mbox{.}}{2020}]%
        {VerWanDoeBac2020cash}
\bibfield{author}{\bibinfo{person}{Diederick Vermetten}, \bibinfo{person}{Hao
  Wang}, \bibinfo{person}{Carola Doerr}, {and} \bibinfo{person}{Thomas
  B{\"a}ck}.} \bibinfo{year}{2020}\natexlab{}.
\newblock \showarticletitle{Integrated vs. Sequential Approaches for Selecting
  and Tuning {CMA-ES} Variants}.
\newblock In \bibinfo{booktitle}{\emph{Proceedings of the Genetic and
  Evolutionary Computation Conference, GECCO 2020}},
  \bibfield{editor}{\bibinfo{person}{Carlos~A. {Coello Coello}}} (Ed.).
  \bibinfo{publisher}{ACM Press}, \bibinfo{address}{New York, NY}.
\newblock
\showISBNx{978-1-4503-7128-5}
\urldef\tempurl%
\url{https://doi.org/10.1145/3377930.3389831}
\showDOI{\tempurl}


\bibitem[\protect\citeauthoryear{Vermetten, Wang, L{\'o}pez-Ib{\'a}{\~n}ez,
  Doerr, and B{\"{a}}ck}{Vermetten et~al\mbox{.}}{2022a}]%
        {zenodo}
\bibfield{author}{\bibinfo{person}{Diederick Vermetten}, \bibinfo{person}{Hao
  Wang}, \bibinfo{person}{Manuel L{\'o}pez-Ib{\'a}{\~n}ez},
  \bibinfo{person}{Carola Doerr}, {and} \bibinfo{person}{Thomas B{\"{a}}ck}.}
  \bibinfo{year}{2022}\natexlab{a}.
\newblock \bibinfo{booktitle}{\emph{Analyzing the Impact of Undersampling on
  the Benchmarking and Configuration of Evolutionary Algorithms -- Dataset}}.
\newblock
\urldef\tempurl%
\url{https://doi.org/10.5281/zenodo.5925410}
\showDOI{\tempurl}


\bibitem[\protect\citeauthoryear{Vermetten, Wang, L{\'o}pez-Ib{\'a}{\~n}ez,
  Doerr, and B{\"{a}}ck}{Vermetten et~al\mbox{.}}{2022b}]%
        {figshare_figures}
\bibfield{author}{\bibinfo{person}{Diederick Vermetten}, \bibinfo{person}{Hao
  Wang}, \bibinfo{person}{Manuel L{\'o}pez-Ib{\'a}{\~n}ez},
  \bibinfo{person}{Carola Doerr}, {and} \bibinfo{person}{Thomas B{\"{a}}ck}.}
  \bibinfo{year}{2022}\natexlab{b}.
\newblock \bibinfo{title}{Performance Variability - Figures}.
\newblock
\newblock
\urldef\tempurl%
\url{https://doi.org/10.6084/m9.figshare.18857486.v1}
\showDOI{\tempurl}


\end{thebibliography}

\providecommand{\MaxMinAntSystem}{{$\cal MAX$--$\cal MIN$} {Ant} {System}}
  \providecommand{\rpackage}[1]{{#1}}
  \providecommand{\softwarepackage}[1]{{#1}}
  \providecommand{\proglang}[1]{{#1}}

\end{document}